\documentclass[10pt,journal,compsoc]{IEEEtran}
\ifCLASSOPTIONcompsoc
  \usepackage[nocompress]{cite}
\else
  \usepackage{cite}
\fi
\ifCLASSINFOpdf
  \usepackage[pdftex]{graphicx}
\else
\fi
\usepackage{amsmath, amsfonts}
\usepackage{algorithm}
\usepackage{algorithmicx}
\usepackage{algpseudocode} 
\usepackage{booktabs}

\usepackage{array}
\usepackage{url}

\usepackage{multirow}
\usepackage[table,xcdraw]{xcolor}

\usepackage{color}
\usepackage{soul}
\usepackage[colorlinks,
            linkcolor=red,      
            anchorcolor=red,  
            citecolor=blue,       
            ]{hyperref}

\usepackage{changes}
\definecolor{mistyrose}{rgb}{1.0, 0.89, 0.88}

\newcommand{\net}{TxST\ }

\definechangesauthor[name=wang, color=blue]{wang}
\definechangesauthor[name=liu, color=blue]{liu}
\definechangesauthor[name=vicky, color=red]{vicky}
\definechangesauthor[name=siu, color=red]{siu}

\begin{document}
%
\title{Name Your Style: An Arbitrary Artist-aware Image Style Transfer}
%
%
%
%

\author{Zhi-Song~Liu$^*$,
        Li-Wen~Wang$^*$,
        Wan-Chi~Siu,~\IEEEmembership{Life~Fellow,~IEEE}
        and~Vicky~Kalogeiton 
\IEEEcompsocitemizethanks{\IEEEcompsocthanksitem Zhi-Song Liu$^*$ (Equal contribution) is the corresponding author, he is with Caritas Institute of Higher Education, Hong Kong. \protect\\
Email: zhisong.liu@connect.polyu.hk
\IEEEcompsocthanksitem Li-Wen Wang$^*$ (Equal contribution) is with The Hong Kong Polytechnic University, Hong Kong. Email: liwen.wang@connect.polyu.hk
\IEEEcompsocthanksitem Professor Wan-Chi Siu is with Caritas Institute of Higher Education and The Hong Kong Polytechnic University, Hong Kong. \protect\\
Email: enwcsiu@polyu.edu.hk
\IEEEcompsocthanksitem Vicky Kalogeiton is with LIX, École Polytechnique, CNRS, IP Paris. Email: vicky.kalogeiton@lix.polytechnique.fr}
}

%
%

\markboth{submitted to IEEE Transactions on Pattern Analysis \& Machine Intelligence}%
{Shell \MakeLowercase{\textit{et al.}}: Bare Demo of IEEEtran.cls for Computer Society Journals}
%



\IEEEtitleabstractindextext{%
\begin{abstract}
Image style transfer has attracted widespread attention in the past few years. Despite its remarkable results, it requires additional style images available as references, making it less flexible and inconvenient. Using text is the most natural way to describe the style. More importantly, text can describe implicit abstract styles, like styles of specific artists or art movements. In this paper, we propose a text-driven image style transfer (TxST) that leverages advanced image-text encoders to control arbitrary style transfer. We introduce a contrastive training strategy to effectively extract style descriptions from the image-text model (i.e., CLIP), which aligns stylization with the text description. To this end, we also propose a novel and efficient attention module that explores cross-attentions to fuse style and content features. Finally, we achieve an arbitrary artist-aware image style transfer to learn and transfer specific artistic characters such as Picasso, oil painting, or a rough sketch. Extensive experiments demonstrate that our approach outperforms the state-of-the-art methods on both image and textual styles. Moreover, it can mimic the styles of one or many artists to achieve attractive results, thus highlighting a promising direction in image style transfer.
\end{abstract}

}

\maketitle

\IEEEdisplaynontitleabstractindextext

%
\IEEEpeerreviewmaketitle

\begin{figure}[t]
	\vskip 0.01in
	\vspace{-1mm}
	\begin{center}
		\centerline{\includegraphics[width=1\columnwidth]{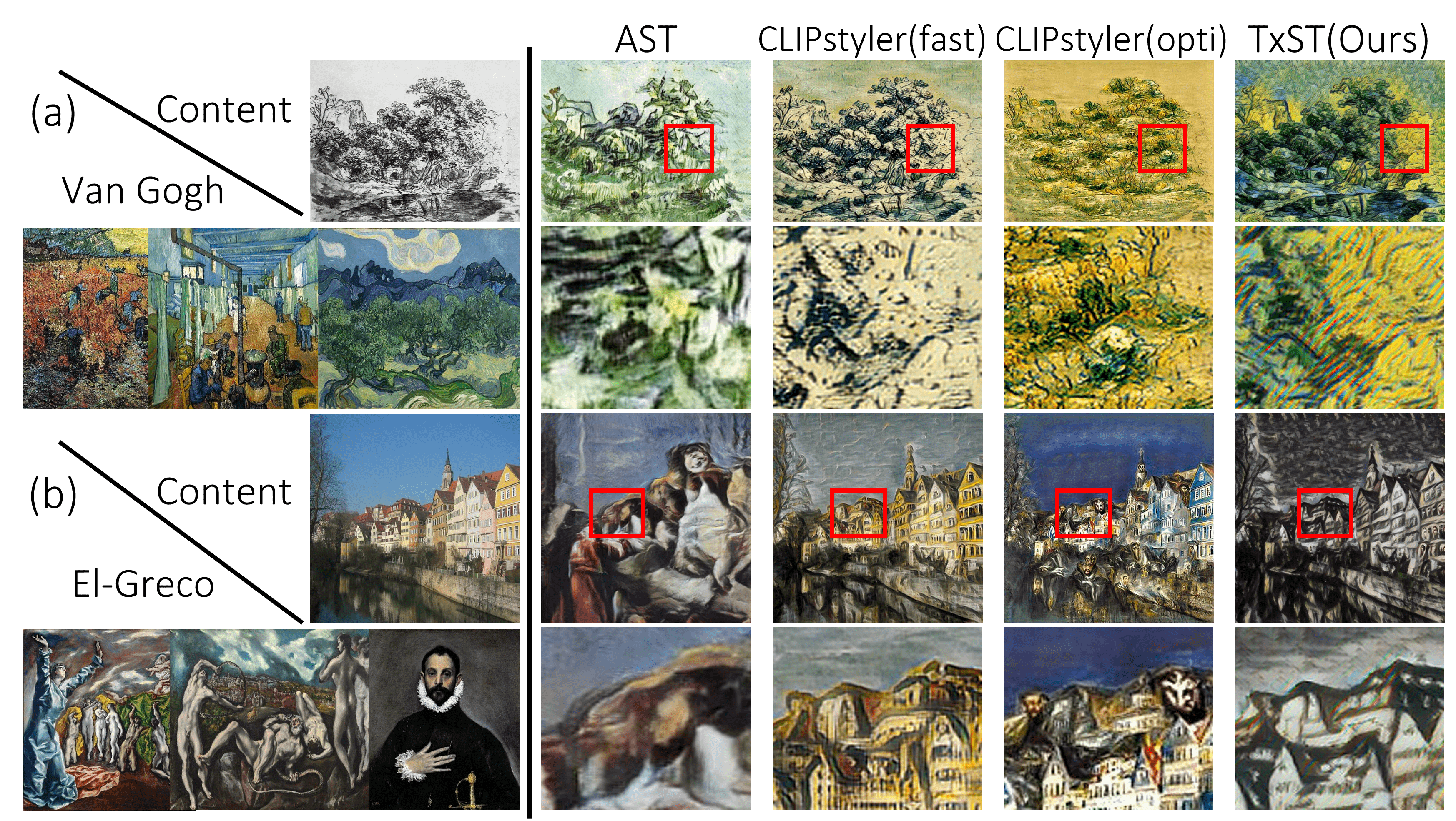}}
		\caption{\small{\textbf{Artist-aware style transfer.} We show two examples of using different methods for style transfer:  (a) AST (style-specific), (b) Clipstyler (fast and opti) and (c) our proposed \net. Example (a) uses a sketch as a content to transfer \textit{Van Gogh} style. Example (b) uses a photo as a content to transfer \textit{El-Greco} style. AST fails to preserve the content information and CLIPstyler cannot transfer the correct styles from the text, such as the faces on the wall produced by CLIPstyler (opti) in (b). Instead, \net successfully  mimics the artists' style (curvature of \textit{Van Gogh}, dark twisted lines of \textit{El-Greco}) while preserving the content cues.
		}}
		\label{fig:teaser}
	\end{center}
	\vskip -0.3in
\end{figure}

\IEEEraisesectionheading{\section{Introduction}\label{sec:introduction}}
\IEEEPARstart{I}{mage} style transfer is a popular topic that aims to apply desired painting style onto an input content image. The transfer model requires the information of \textit{``what content"} in the input image and \textit{``which painting style"} to be used. Conventional style transfer methods require a content image accompanied by a style image to provide the content and style information. However, people have specific aesthetic needs. Usually, it is inconvenient to find a proper style image that perfectly matches one's requirements. Text or language is a natural interface to describe which style is preferred. Instead of using a style image, using text to describe style preference is easier to obtain and more adjustable. 
Furthermore, achieving perceptually pleasing artist-aware stylization typically requires learning from collections of arts, as one reference image is not representative enough. 
%
Figure~\ref{fig:teaser} shows the artist-aware stylization (Van Gogh and El-Greco) on two examples, a sketch\footnote{\textit{Landscape Sketch with a Lake} drawn by Mark\'{o}, K\'{a}roly (1791–1860)} and a photo. 
For each example, we display four stylizations: 
(1) style-specific optimization (AST~\cite{ast}), (2,3) text-driven optimization (CLIPstyler~\cite{language_2}, and text-aware stylization (Ours, \net). Note there are two versions of CLIPstyler. CLIPstyler(opti) requires real-time optimization on each content and each text. CLIPstyler(fast) requires real-time optimization on each text. Hence, both CLIPstyler and AST are time-consuming. We observe that AST grasps the style from the artist's work, but it does not preserve the content. CLIPstyler(slow) fails to learn fine styles and colors. CLIPstyler(opti) also fails to learn the most representative style but instead, it pastes specific patterns, like the face on the wall in Figure~\ref{fig:teaser}(b). 
In contrast, \net takes arbitrary texts as input\footnote{\net can also take style images as input for style transfer, as shown in the experiments.} for real-time style transfer; this enables faithfully  mimicing the artists' style (the curvature of \textit{Van Gogh} and dark twisted lines of \textit{El-Greco}) while also preserving the content information.

In this work, we aim to learn arbitrary artist-aware image style transfer, which transfers the painting styles of any artists to the target image using texts and/or images. Previous studies on universal image style transfer~\cite{LT,sanet,ast} limit their applications to using reference images as style indicator that are less creative or flexible. As shown in Figure~\ref{fig:teaser}, AST~\cite{ast}, one of the state-of-the-art artist style transfer methods, learns a dedicated model for one specific style, which usually overemphasizes the styles from artists that the output loses content information. On the contrary, \net can use the text \textit{Van Gogh} to mimic the distinctive painting features (e.g., curvature) onto the content image. Text-driven style transfer has been studied by~\cite{language_1} and \cite{language_2}, and shown some promising results using a simple text prompt. However, they either require costly data labelling and collection, or require online optimization for each content and each style (as CLIPstyler(fast) and CLIPstyler(opti) in Figure~\ref{fig:teaser}). Our proposed \net overcomes these two problems and achieves much better and more efficient stylization.  

Previous works, like CLIPstyler, have been devoted to implementing text-driven style transfer. Our proposed \net explicitly adopts this idea to achieve artist-aware style transfer. That is, we try to find the hidden space where the global distance of different artworks (different artists) can be maximized, while the same artworks (same artists) can be minimized. The process of our proposed \net is shown in Figure~\ref{fig:method_compare}. We randomly sample two style images ($I_s^1$ and $I_s^2$) from two artists, then we minimize the distance between the target texts and generated images ($\Delta T_1$ and $\Delta T_2$), while maximizing the distance between two generated images $\Delta D$. It can resonate with the classic Fisher Linear Discriminant~\cite{fda} to distribute features into groups. Specifically, given artists' names known as a prior, we project features from different artworks onto the CLIP space for classification. 

In this work, we empirically analyze the co-linearity between artists and paintings on the CLIP space to demonstrate the reasonableness and effectiveness of text-driven style transfer. Our experiments on both image- and text-driven style transfer show that the proposed approach outperforms other state-of-the-art quantitatively and qualitatively.
The key contributions of our proposed artist-aware image style transfer can be summarized as follows.

\begin{itemize}
	\item To achieve text-driven image style transfer, we propose to embed the task-agnostic image-text model, i.e., CLIP, into our network. This enables our network to obtain style preference from images or text descriptions, making the image style transfer more interactive. (Sections~\ref{sec:preliminaries},\ref{sub:textstyle}) 
	\item We propose a polynomial attention module to estimate polynomial cross-correlations between content and style features. To restore style preference in the spatial domain, we propose a novel positional mapper that expands the 1-D CLIP code into 2-D space for position awareness. Based on the restored style preference and cross-correlations, the content feature is then projected with target style preference, which significantly improves the effectiveness of the style transfer (Section~\ref{sub:txst}). 
	\item We propose to use Constrastive training strategy (Sections~\ref{sec:preliminaries},~\ref{sub:txst}) to learn art collection awareness, equivalent to a self-supervised classification. This studies the inter-class similarity of different artists without explicit label guidance and data collection.
	\item We conduct extensive experiments on both general and artist-aware style transfer using both style images and texts. We show that our proposed \net outperforms the state of the art using both quantitative and qualitative evaluations. 
	To demonstrate the generability and applicability of \net to other tasks, we also conduct studies on multiple style transfer and high-resolution style transfer. 
\end{itemize}






\section{Related Work}
\label{sec:relwork}
Here, we introduce three related topics: arbitrary and domain-aware style transfer and text-driven image editing.

\noindent \textbf{Arbitrary style transfer.} It can be categorized into two groups: (1) style-aware optimization~\cite{Gatys,Ulyanov,wct,strotss} and (2) universal style transfer~\cite{adain,LT,avatar,art-net}. 

The former explicitly studies the statistic correlations between content and style images in the deep feature space. Gatys et al.~\cite{Gatys} propose the first flexible iterative optimization approach based on a pre-trained VGG19 network~\cite{VGG}. It can achieve arbitrary style transfer, but the forward and backward passes are time-consuming. STROTSS~\cite{strotss} replaces the l-1 style loss to Relaxed EMD distance to achieve higher visual quality. WCT~\cite{wct} studies the second-order correlations via whitening and coloring transforms without costly optimization. To further reduce the computation of WCT, LT~\cite{LT} uses linear transformations to approximate the matrix decomposition, resulting in real-time style transfer. Another direction on statistic optimization is to use attention mechanism~\cite{attention} to explore spatial  correlations. SANet~\cite{sanet} matches the content and style statistics via cross attention. AdaAttN~\cite{adaattn} further explores the second-order attention to preserve more content information without losing style patterns. Most recently, Artflow~\cite{artflow} and VAEST~\cite{vaest} investigate the normalization flow~\cite{flow} and VAE~\cite{vae} to fuse style and content images, which also produce very promising results.

The latter focuses on zero- or first-order statistics for real-time style transfer. AdaIN~\cite{adain} proposes Adaptive instance normalization to shift the deep content feature to the style space. Avatar-net~\cite{avatar} proposes to combine AdaIN~\cite{adain} and style swap~\cite{swap} to enhance the stylization in a coarse-to-fine manner. ReReVST~\cite{rerevst} follows Avatar-net~\cite{avatar} to further develop inter-channel feature adjustment for both image and video style transfer. To handle high-resolution images, ArtNet~\cite{art-net} proposes Zero-channel Pruning to reduce model complexity. Wang et al.~\cite{wang} propose Collaborative Distillation to compress the pre-trained network (e.g., AdaIN) for fast computation.

\noindent\textbf{Domain-aware style transfer.} The problem of arbitrary style transfer methods is that they learn style statistics from one reference image, which is not representative enough for the desired pattern. Domain-aware style transfer is to learn the robust statistics from many similar images so that it can transfer the most distinctive styles to the content images. The straightforward approach is to collect a number of desirable images to train a specific model for style transfer. For example, AST~\cite{ast} proposes an artist-aware style transfer to achieve art stylization. In other words, they collect specific artworks as style references, e.g., Van Gogh, to train a network for specific style transfer. DualAST~\cite{dualast} follows this idea and proposes a more flexible deep network to balance both artwork style and artist style via Style-Control Block. Inspired by the study on the disentangled representation~\cite{infogan}, domain-aware style transfer can also be regarded as an image2image problem where the image is factorized into class-related (style) and class-independent components (content). Kotovenko et al.~\cite{Kotovenko} propose a double-GAN model to interpret the encoder and decoder for feature decomposition and stylization via two discriminators. ~\cite{munit,drit,drit_plus} further investigate cross-cycle domain consistency for style transfer and other applications. Most recently, there are some new developments on transformer based style transfer~\cite{vision_transformer_1,vision_transformer_2}.

\noindent\textbf{Text-driven image style transfer.}
Style transfer is a subjective topic, that is, different people may have different preferences for stylization.
Using style images as references may not be as sufficiently good as texts can describe styles in a more abstract and aesthetic manner.
The success of CLIP~\cite{clip}, VQVAE~\cite{vqvae} and multimodality~\cite{multi_1,multi_2,multi_3} show that text and image can be related via the shared projection space. Some pioneer works on image editing~\cite{dall-e,stylegannada,clip_draw_1,clip_draw_2,styleclip,projectedGAN2021,huang2021multimodal,liu2021control,VisualConceptVocabulary} and video understanding~\cite{univl,clip2video,clip4clip} show that language can be used for user guided applications. \cite{VisualConceptVocabulary} constructs open-ended vocabularies to flexibly recompose the visual content in the latent spaces of GANs. Most recently, ~\cite{language_1} uses the CLIP as the condition for style transfer that increases the cross-correlation between the output and text description for text-guided style transfer. ~\cite{language_2} further develops this idea by using both global and patch CLIP losses to generate high-resolution stylized images. 
Instead, in this work, we rely on the co-linearity between texts and images and  train our model using contrastive similarity. Thus, both texts and images with the same style patterns can bring the stylization close to each other; otherwise, they push each other away.

\section{Preliminaries}
\label{sec:preliminaries}

At the core of our proposed TeXt-driven image STyle transfer \net, there are two key components: CLIP~\cite{clip} and Contrastive representation learning~\cite{simclr,simclr2}. In the following, we discuss how CLIP has been employed in this context in the past and how we use Contrastive representation learning to embed CLIP for artist-aware style transfer.

\begin{figure}[t]
	\begin{center}
		\centerline{\includegraphics[width=1\columnwidth]{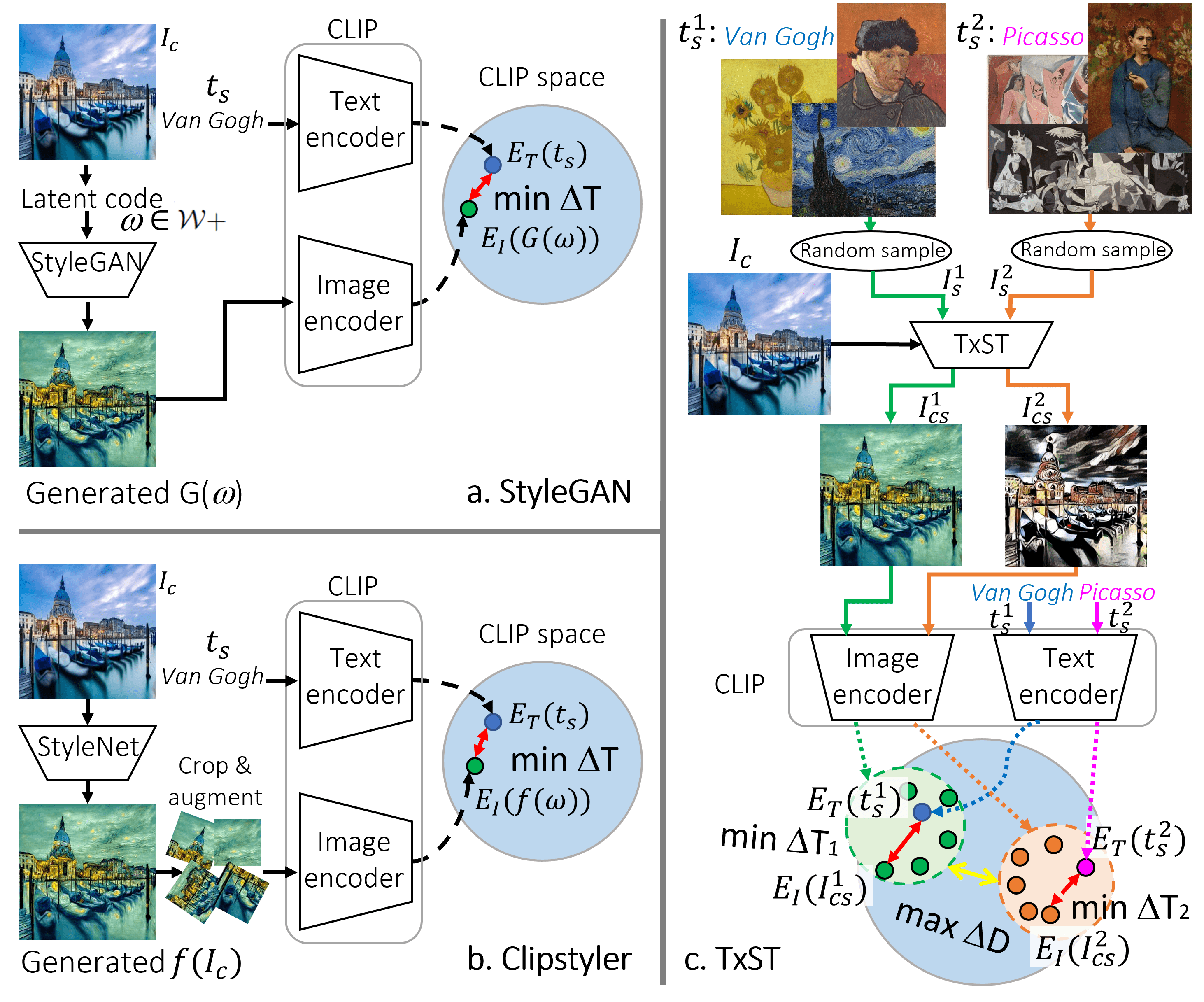}}
		\caption{\small{\textbf{Comparison among different CLIP based image manipulation.} We show the overall training processes of a) StyleCLIP, b) Clipstyler and c) our proposed \net for comparison. They all use CLIP image encoder $E_I$ and text encoder $E_T$ to project images and texts onto the CLIP space for distance $\Delta T$ measurement. For a), it sample latent code $\omega$ from the pre-define latent space $\mathcal{W}+$ as input to StyleGAN for generation. For b), it uses source image $I_c$ as input to the StyleNet to obtain $f(I_c)$. For c), it samples two random style images $I_s^1$ and $I_s^2$ for style transfer. By using the contrastive training loss, it minimizes the intra-class distances (text-image) $\Delta T_1$ and $\Delta T_2$, as well as maximizes the inter-class distance (image-image) $\Delta D$.
		}}
		\label{fig:method_compare}
	\end{center}
\end{figure}

\subsection{styleCLIP}
StyleCLIP~\cite{styleclip} uses the text description only for arbitrary facial image editing, like adding mustache or glasses. The idea is to combine the generative power of styleGAN~\cite{stylegan,stylegan2} with the semantic abstraction of CLIP~\cite{clip} to discover latent space attribute editing. StyleCLIP provides three options for achieving text-guided manipulation. The first two approaches directly optimize the CLIP-space distance between text and image, with dedicated models for desirable changes. Similar concepts are further developed by CLIPstyler~\cite{language_2}, which takes augmented texts with the same meaning to iteratively optimize the network to produce reasonable stylizations. Figure~\ref{fig:method_compare} shows both processes of StyleCLIP and CLIPstyler. Given the target text $t_s$, their common goal is to optimize the distance $\Delta T$ to the generated image in the Clip space. Their difference is that StyleCLIP learns the generated image $G(\omega)$ from the latent code $\omega$, while CLIPstyler learns it $f(I_c)$ from the source image $I_c$.
The third approach disentangles the feature representation in the latent space to find the latent-code entries that induce both the image-space change and the global direction of the two text descriptions (source and target). 

\subsection{simCLR}
\label{sub:simclr}

For text-driven style transfer, there is one key issue to address: `Is text rich enough to represent the key features?'
Text can explicitly describe the desirable styles, like \textit{stormy night}, \textit{colorful oil painting}, but usually it represents more abstract styles, like the name of artists (\textit{Van Gogh}) or the name of art movements (\textit{Impressionism}). These do not describe the visual styles. For StyleCLIP and CLIPstyler, they reply on the power of CLIP that it can capture the co-linearity between texts and images. When the training data are not sufficiently numerous or the training time is not enough, the quality of the generated images may become worse. The study on Contrastive representations learning~\cite{simclr,simclr2} provides an alternative to text-driven style transfer. Instead of learning dedicated mapping models for image generation, given priors of style texts, we learn to classify them in the CLIP space, hence arbitrary text-driven image style transfer can be achieved. In other words, we learn to maximize the prior estimation of text distribution on the CLIP space. As shown in Figure~\ref{fig:method_compare}, by finding the principal direction, where the text features are projected on, that maximizing the distance between different texts, we can learn the most distinct feature representation for each text. In the meantime, we use CLIP to minimize the distance between paired images and texts. We describe this mathematically as,

\begin{small}
	\begin{equation} \tag*{(1)}
	\begin{matrix}
	\!\begin{aligned}
	L_{\text{MC}} &= min\Delta T - max \Delta D \\
	&=min\sum_{i=1, j=1}^{i=N,j=C} \big(E_I(I_{cs}^{i,j}),~E_T(t_{s}^j)\big) +\\ &~~~~min\sum_{i=1, j=1, k\ne i}^{i,j=N,j=C} \big(E_I(I_{cs}^{i,j}),~E_I(I_{cs}^{k,j})\big)
	\label{eq:Equation1}
	\end{aligned}
	\end{matrix} 
	\end{equation}
\end{small}

\noindent Equation~\ref{eq:Equation1} minimizes the intra-class distance $\Delta T$ between paired images and texts, while maximizing the inter-class distance $\Delta D$ between images from different classes. The maximization of inter-class distance can further be rewritten as minimizing the inter-class distance between images from the same classes.

\section{Approach}

\subsection{Text for Style Transfer}
\label{sub:textstyle}
Painting style, or \textit{painting language}, represents the painting tastes of artists. It is a high-level abstract representation of the images. Previous style transfer methods~\cite{adain,art-net,sanet} use a referenced image to describe the desired painting style. The core idea of our approach is to describe painting style by means of texts, because of its convenience and flexibility. The benefits of the text model are twofold: (1) high-level style representation that avoids ambiguity caused by the image content; and (2) flexible adjustment with no effort for searching proper references. Recently developed text models, like CLIP\cite{clip}, 
motivate us to ask \textit{``Can text models be aware of different style representations?"} 

\begin{figure}[!t]
\centering
\includegraphics[width=\linewidth]{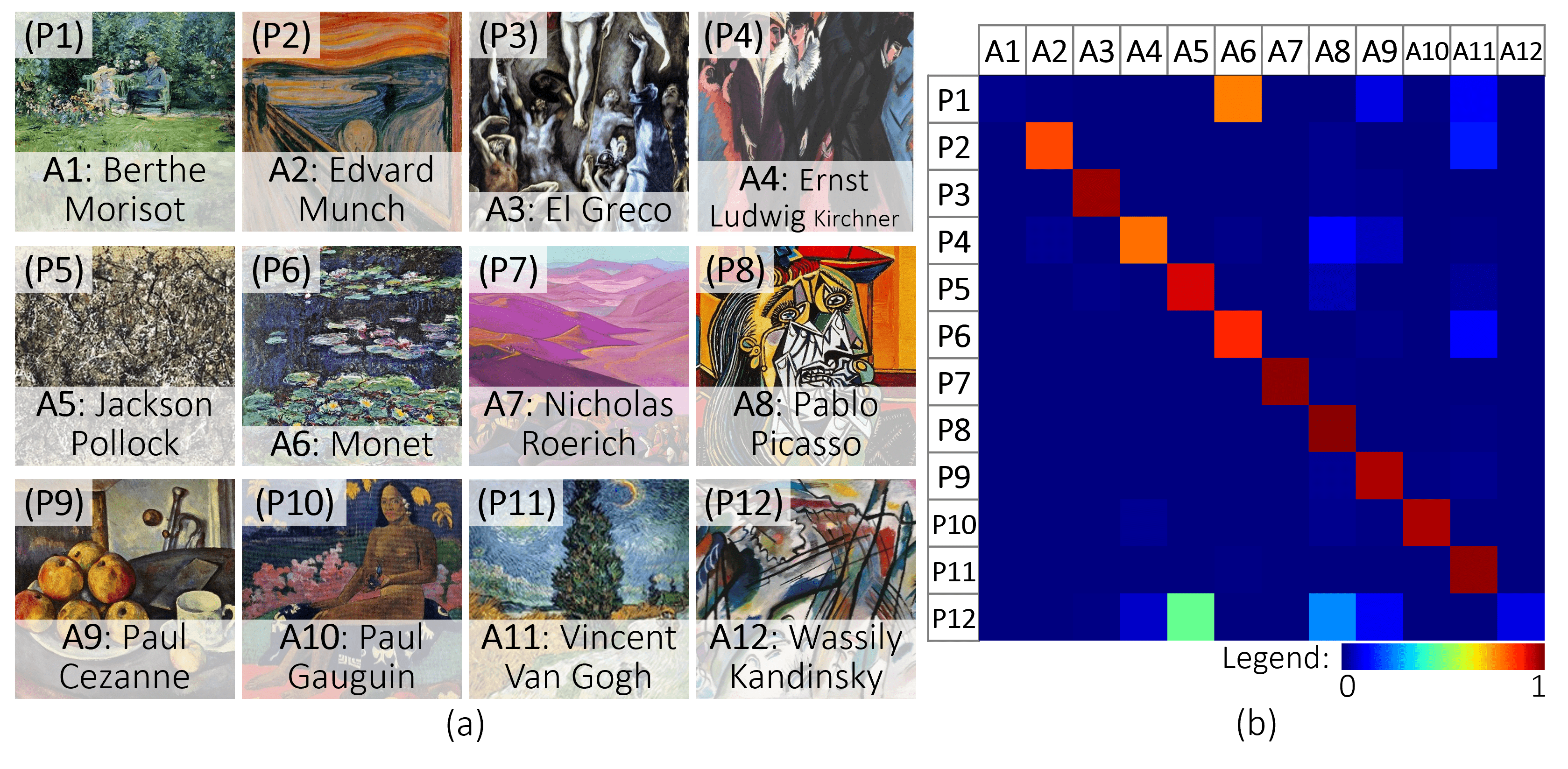}
\caption{\small{\textbf{Correlation between artists and paintings.}(a) A set of Artist-Painting paired samples from WikiArt dataset \cite{WikiArt}. The artists (Abbr.: \textbf{A}) are represented by the means of text, and their paintings (Abbr.: \textbf{P}) are in the format of color images. Different artists and their paints are given different index numbers for clear visualisation. (b) Feature relationship between artists and paintings. Features of the artists and paintings are extracted by the CLIP\cite{clip} with language and visual portions, respectively. The horizontal and vertical axes are the artists and paintings respectively. The significant larger values on the diagonal elements suggests that the features from CLIP\cite{clip} model are aware of the high-level painting style of different artist.}}
\label{fig_ana_wiki}
\end{figure}

Following the discussion of Section 3, we believe that text and image can work interchangeably in the CLIP space for style transfer. In other words, text and image are co-linear in the CLIP space and hence, they can both be used as style indicators. From the aspect of the style description, this co-linearity also exists between artists and their paintings, making it possible to realize artist-aware style transfer. To verify this, we make an analysis between images and style queries using a pre-trained CLIP model\footnote{If there is no special instruction, the language model refers to CLIP~\cite{clip} with the backbone setting of "ViT-B/32", implemented by \url{https://github.com/openai/CLIP}.}. 
Initially, we collect a set of Artist-Painting pairs from WikiArt dataset~\cite{WikiArt}, as shown in Figure~\ref{fig_ana_wiki}(a). To investigate the correlation between artists and paintings, the names of the artist are firstly tokenized and encoded to extract the feature $E_{T}\in\mathbb{R}^{512}$. Next, we encode the paintings for the image feature $E_{I}\in\mathbb{R}^{512}$. The correlation is then calculated by the dot product between the text $E_{T}(t_s)$ and image features $E_{I}(I_s)$. For each painting $I_s^i$, we use the softmax function to find the probability score $s$ for different artists $t_s^j$, as follows:

\begin{small}
\begin{equation} \tag*{(2)}
\label{eq_ana_clip}
s\left( i,j \right) =\frac{\exp \left( E_{I}(I_s^i)\cdot E_{T}(t_s^j) \right)}{\sum\nolimits_{j=0}^{j=C}{\exp \left( E_{I}(I_s^i)\cdot E_{T}(t_s^j) \right)}}
\end{equation}
\end{small}

\noindent where $C$ is the number of artists. A larger score $s\left( i,j \right)$ means more likely the painting $I_s^i$ is drawn by the artist $t_s^j$.
The relationship among scores is shown in Figure~\ref{fig_ana_wiki}(b). It is clear that there are significantly larger values on the diagonal. Intuitively, we observe a strong correlation between the artists and their paintings in the CLIP feature space\footnote{There are two outliers in the figure, i.e., ``\textbf{P}1-\textbf{A}6" and ``\textbf{P}12-\textbf{A}5". They indicate that the painting \textbf{P}1 by \textit{Berthe Morisot} has the highest score to the painting of artist \textit{Monet}(\textbf{A}6). The reason is that, both artists lived in the same country (France), and the artworks were created during the same artistic period (both were born in the 1840s) which we call it the Impressionism movement. For another outlier ``\textbf{P}12-\textbf{A}5", the painting from \textit{Wassily Kandinsky} has similar styles to \textit{Jackson Pollock}. Both paintings \textbf{P}5 and \textbf{P}12 in Figure~\ref{fig_ana_wiki}(a) use geometric symbols in complex and abstract forms which are difficult to distinguish.}.

\subsection{Proposed text-driven image style transfer (TxST)}
\label{sub:txst}
\begin{figure}[t]
	\begin{center}
		\centerline{\includegraphics[width=1\columnwidth]{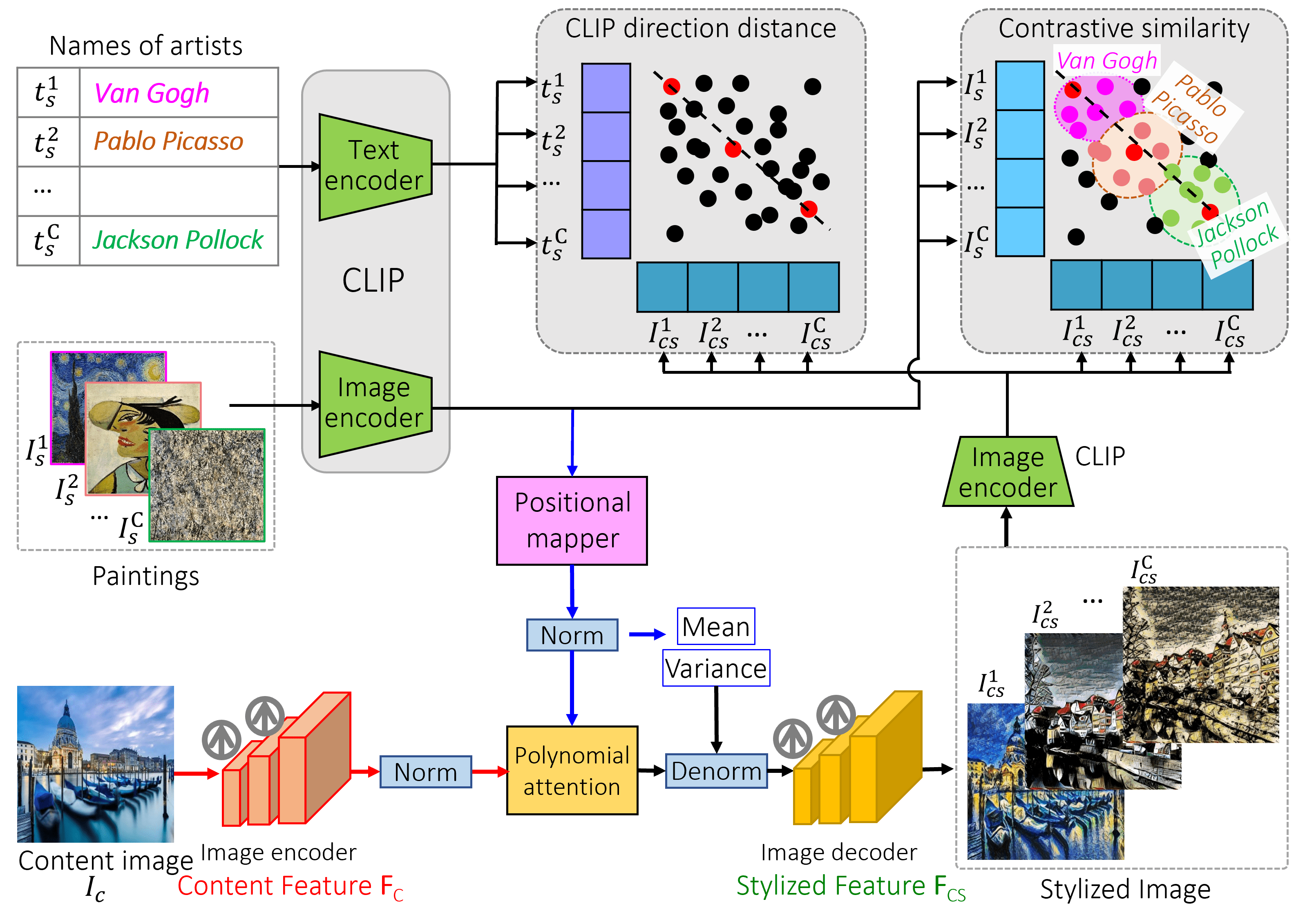}}
		\caption{\small{\textbf{Training process of the proposed \net.} We show the overall training processes of our proposed \net. Given content images, artists' names and paintings, we process them to maximize the style's variance, as well as maximize the similarity of the same style.
		}}
		\label{fig:network}
	\end{center}
\end{figure}

Figure~\ref{fig_ana_wiki} highlights that CLIP-based text and image features are colinear to each other for style transfer. \footnote{We also analyze the text-image relationship on textures and edges. The results are shown in the supplementary material.}
Based on this finding, here, we introduce the architecture of our proposed \net for artist-aware image style transfer. Figure~\ref{fig:network} illustrates it.


Given content images $I_c$ and desirable styles, artist's name $t_s$ and corresponding painting $I_s$, \net outputs the stylized image $I_{cs}$. In the training stage, we use CLIP to maximize the variance of different styles by preserving the co-linearity between texts and style images. The network computes the contrastive similarity to minimize the inter-distance of the same styles. At test time, both text and image can be input as style indicators for arbitrary style transfer. 

\net consists of four parts: Image encoder, Image decoder, CLIP (text and image encoders), Positional mapper, and Polynomial attention. The structure of the image encoder is of the same structure as VGG-19~\cite{VGG}, discarding the fully connected layers. The image decoder is symmetric to the image encoder, with gradual upsampling feature maps towards final stylized images. We use CLIP to encode the paired texts ($t_s^i, i=1,2,..,C$, where $C$ is the number of artists) and images ($I_s^i, i=1,2,..,N$, where $N$ is the number of paintings) to obtain corresponding text features $E_T(t_s^i)$ and image features $E_I(I_s^i)$, respectively. 

\net uses CLIP features as the style condition to edit the content features. As shown in Figure~\ref{fig:network}, we process them by using Positional mapper to project it from 1-D vector to 2-D map, as $F_s=M\big(E_I(I_s^i)\big)$. Then, the proposed Polynomial Attention module $P_R$ is used to fuse content and style features to obtain stylized features $F_{cs}=\sigma(F_s)\times P_R\big(\Bar{F_s},\Bar{F_c}\big)+\mu(F_s)$, where $\Bar{F_s},\Bar{F_c}$ are normalized content and style features, $\mu(F_s), \sigma(F_s)$ are mean and variance of the style features. Below, we introduce the proposed Positional mapper and Polynomial Attention module in details.

\begin{figure}[t]
	\begin{center}
		\centerline{\includegraphics[width=\columnwidth]{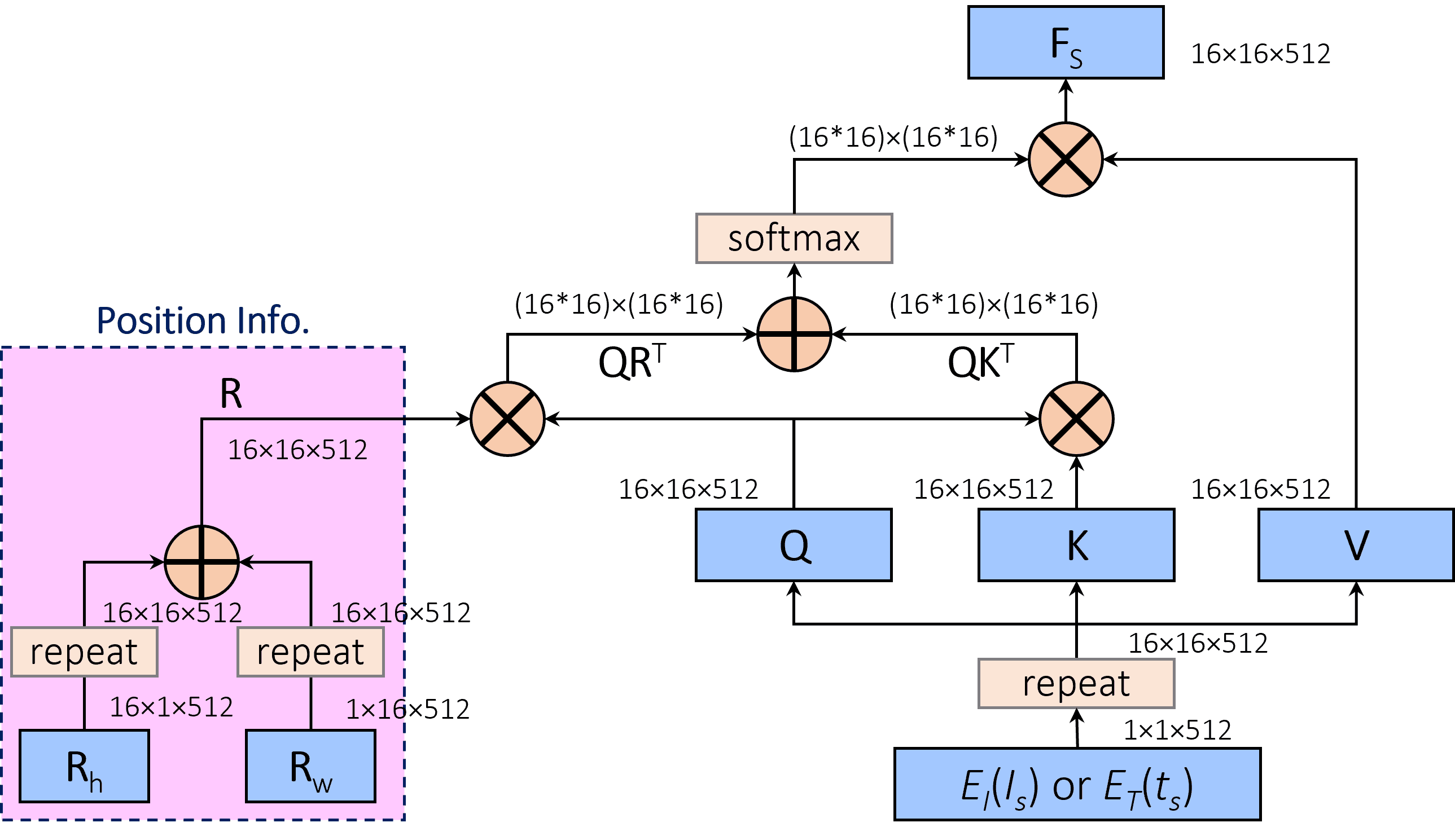}}
		\caption{\small{\textbf{Structure of the Position mapper.} $\textbf{R}_h$ and $\textbf{R}_w$ are two learnable position encodings for horizontal and vertical directions. Relative position encoding $\textbf{R}$ is sum of the  $\textbf{R}_h$ (repeat 16 times along width dimension) and $\textbf{R}_w$ (repeat 16 times along width dimension). The input style vector can come from the image $E_{I}(I_s)$ or text $E_{t}(T_s)$. The repeated vector is then processed by three 1-by-1 linear projection layers to form a common Self-Attention (SA) \cite{non-local} with query $\textbf{Q}$, key $\textbf{K}$ and value $\textbf{V}$. The relative position encoding $\textbf{R}$ is multiplied with the query $\textbf{Q}$ and attends the SA process. Finally the position mapper produces a position-aware style feature $\textbf{F}_{s}$. The dimensions in the figure are in the order of $height\times width \times channels$.    
		}}
		\label{fig:position}
	\end{center}
\end{figure}

\noindent $\bullet$ \textbf{The positional mapper.} CLIP projects the image and text into 1-D vectors as $E_{I}(I_s)\in\mathbb{R}^{512}$ and $E_{t}(T_s)\in\mathbb{R}^{512}$, which will result in lossing the spatial information for style transfer. Instead, to make style transfer \textit{position aware}, we propose a Positional mapper to embed position information into the 1-D style vector, as shown in Figure~\ref{fig:position}. Following the study on relative position encodings~\cite{botnet}, we design our Positional mapper structure to introduce position information into the style vector. As shown in Figure~\ref{fig:position}, it first defines two learnable relative position encodings $\textbf{R}_h\in\mathbb{R}^{16\times1\times512}$ and $\textbf{R}_w\in\mathbb{R}^{1\times16\times512}$ for height and width, respectively. The two position encodings are repeated to the same size such that they add together for a relative position coding map $\textbf{R}\in\mathbb{R}^{16\times16\times512}$. The relative position map benefits the network to be aware of different locations on the spatial plane. Next, the position coding map attends the Self-Attention (SA)~\cite{non-local} process, as shown in Figure~\ref{fig:position}. During the SA process, the position coding map $\textbf{R}$ brings in relative distances between features into the style features at different locations, which expands the 1-D style vector into the 2-D plane $\textbf{F}_{s}\in\mathbb{R}^{16\times16\times512}$ at the same time. By using Positional mapper, we compute style transfer in the spatial domain to restore more complex visual patterns.

\noindent $\bullet$ \textbf{Polynomial attention module.} Using attention for style transfer has been studied in~\cite{sanet,adaattn}. The basic idea is to project style features onto the content features via cross attention. We further develop this idea as polynomial attention. In other words, we compute $i^{th}$ degree polynomial in style features to fit the non-linear correlations to the content. The computation of polynomial attention is shown in Figure~\ref{fig:poly_attn}. 
Specifically, we use multiple cross attention blocks to compute multiple polynomial attention between normalized content ($\Bar{F}_c$) and $i^{th}$ order of style ($\Bar{F}_s^i$) features. Finally, we use the mean and variance of style features ($\mu$ and $\sigma$) to denormalize the output. Mathematically, we have: 

\begin{figure}[t]
	\begin{center}
		\centerline{\includegraphics[width=1\columnwidth]{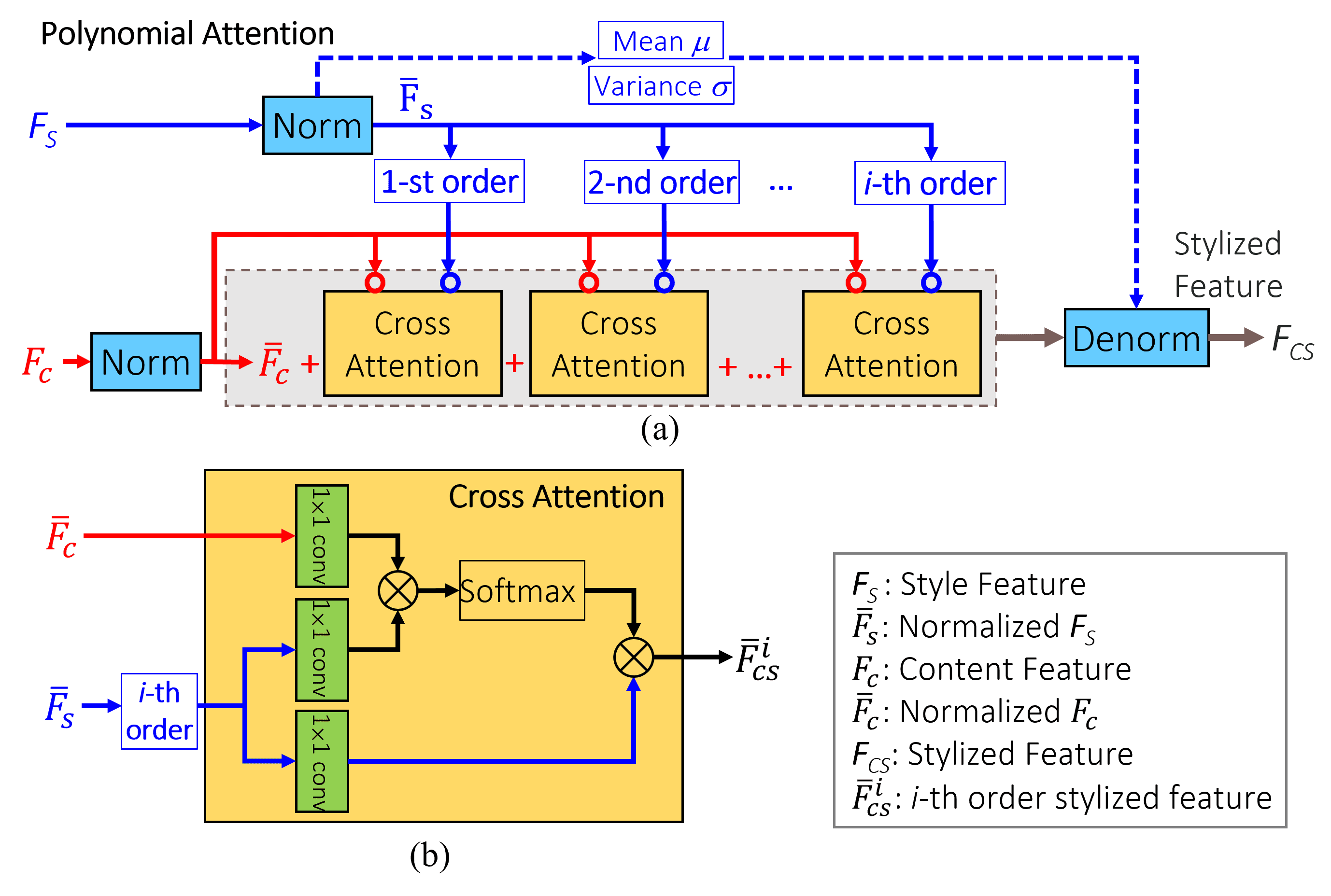}}
		\caption{\small{\textbf{The structure of the proposed Polynomial attention module.} (a) Overall structure of the Polynomial attention module. (b) Structure of the cross attention operation. The Polynomial attention uses multiple cross attention (yellow boxes) to compute correlations between normalized content and $i^{th}$ order of style features. Then it denormalizes the output feature using style mean and variance to obtain the final stylized features.
		}}
		\label{fig:poly_attn}
	\end{center}
\end{figure}

\begin{small}
	\begin{equation} \tag*{(3)}
	\begin{matrix}
    \!\begin{aligned}
	& F_{cs} = \sigma(F_s)\cdot P_R\big(\Bar{F}_s,\Bar{F}_c\big)+\mu(F_s)~~~~where\\
	& P_R\big(\Bar{F}_s,\Bar{F}_c\big) = \Bar{F}_c + \sum_{i=1} ^{i=R} Softmax\left(\frac{Q[\Bar{F}_c]\times K[\Bar{F}_s^i]}{\sqrt{d}}\right)V[\Bar{F}_s^i]
	\label{eq:Equation3}
	\end{aligned}
	\end{matrix} 
	\end{equation}
\end{small}

\noindent where $R$ is the highest order of style estimation, $d$ is the dimension of the feature maps. $Q$, $K$ and $V$ are $1\times1$ convolution for query, key and value. The study on the polynomial attention has the same spirit as polynomial regression estimation. The merits of it are twofold: (1) it learns higher-order correlations between content and style images for better approximation, and (2) it adapts to the larger variance of the style features for unbiased prediction.

\subsection{Loss functions} 
\label{sub:losses}

\noindent $\bullet$  \textbf{Directional CLIP loss.} To guide the content image following the semantic of the target text (artist's name), we use directional CLIP loss~\cite{stylegannada,language_2} to align the CLIP-space direction between the text-image pairs of target $t_s$ and output $I_{cs}$. It is defined as:

\begin{small}
	\begin{equation} \tag*{(4)}
	\begin{matrix}
    \!\begin{aligned}
	& \Delta T=E_T(t_s)-E_T(t_o)\\
	& \Delta I=E_I(I_{cs})-E_I(I_c)\\
	& L_{clip}=1-\frac{\Delta I \cdot \Delta T}{|\Delta I| |\Delta T|}
	\label{eq:clip}
	\end{aligned}
	\end{matrix} 
	\end{equation}
\end{small}

\noindent where $t_o$ is the source text for the content image. We define it as \textit{Photo} following the design in ~\cite{stylegannada}. Compared to original CLIP loss~\cite{clip}, it can stabilize the optimization process and produce results with better quality.

\noindent $\bullet$  \textbf{Contrastive similarity loss.} In order to encourage image features to be correlated with the target style and uncorrelated with other styles, we use Contrastive similarity loss~\cite{simclr} to minimize the loss. Given the $i^{th}$ stylized image feature $\nu_i$ and N-1 other $\nu_j$ from the same batch: $\nu_1, \nu_2, \nu_{i-1}, \nu_{i+1}, ..., \nu_N$, we have

\begin{small}
	\begin{equation} \tag*{(5)}
	\begin{matrix}
    \!\begin{aligned}
	& L_{sim} = -log\frac{exp\big(S(\nu_i^I, \nu_j^I)/\tau\big)}{\sum_{j=1, j\neq i}^N exp\big(S(\nu_i^I, \nu_j^I)/\tau)} + \\
	&~~~~~~~~~ -log\frac{exp\big(S(\nu_i^t, \nu_j^t)/\tau\big)}{\sum_{j=1, j\neq i}^N exp\big(S(\nu_i^t, \nu_j^t)/\tau)}
	\label{eq:simclr}
	\end{aligned}
	\end{matrix}
	\end{equation}
\end{small}

\noindent where $\tau$ is the temperature factor, $S$ is the cosine similarity, $\nu_i^I$ is the stylized results using reference image $I$ and $\nu_i^t$ is the stylized results using reference text $t$. Note that in order to distinguish from the CLIP vector $E_I(I_{cs})$, we use symbol $\nu$ to represent the CLIP feature computed after Vision transformer~\cite{clip} and before the final projection layer. In Equation~\ref{eq:simclr}, there are two terms for the loss: The first one is to compute the feature similarity between stylized results obtained by different style images ($\nu_i^I$ and $\nu_j^I$); The second one is to compute the feature similarity between stylized results obtained by different style texts ($\nu_i^I$ and $\nu_j^t$). As shown in Figure~\ref{fig:network}, by using CLIP loss, we can find the co-linearity between style texts and output images (the dash line drawn on red dots), and by using the Contrastive similarity loss, we can minimize the feature distance between style images and output images (pink/yellow/green dots around red dots). Therefore, we guide the model to minimize the intra-class distance between different art collections painted by the same painter, as well as maximizing the inter-class distance between different painters.

\noindent $\bullet$  \textbf{CLIP feature loss.} Studied by Johanson et al.~\cite{Johnson}, using pre-trained VGG to extract VGG features for content and style images can minimize $L_2$-norm  distance between them to encourage stylized output with visual patterns close to the style images. Inspired by it, we propose to use the CLIP feature loss to guide the stylization close to the style feature representation. We define it as:

\begin{small}
	\begin{equation} \tag*{(6)}
	L_{clip_f} = \frac{1}{N}\Vert\nu_i - \nu_j\Vert_{2}^{2}
	\label{eq:clip_feat}
	\end{equation}
\end{small}

\noindent $\bullet$  \textbf{Content and style feature loss.} Following the existing style transfer methods~\cite{adaattn,sanet,vaest}, we employ content and style feature loss by using the pre-trained VGG network to minimize the distance in the deep feature space as:

\begin{small}
\begin{equation} \tag*{(7)}
	\begin{matrix}
    \!\begin{aligned}
	& L_{sty} = \sum_i^4 \Vert \mu\big(W_s^i(I_{cs})\big)-\mu\big(W_s^i(I_s)\big)\Vert_{1}^{1} \\
	&~~~~~~+ \sum_i^4 \Vert\sigma\big(W_s^i(I_{cs})\big)-\sigma\big(W_s^i(I_s)\big)\Vert_{1}^{1} \\
	& L_{con} = \sum_i^2 \Vert W_c^i(I_{cs})-W_c^i(I_c)\Vert_{1}^{1} 
	\label{eq:con_sty}
	\end{aligned}
	\end{matrix}
	\end{equation}
\end{small}

\begin{algorithm}
\caption{Training procedure for \net}
\begin{algorithmic}[1]
\Require 
\State Network Initialization, fix CLIP and VGG;
\State batch size N, input content image $I_c$, artist's name $t_s$ and painting $I_s$;
\Ensure $I_{cs}$
\For{sampled minibatch $\{I_c^i\}_{i=1}^N$}
    \For{All $i\in \{1,...,N\}$} 
    \State randomly draw two paintings from the same artist ($t_s^i$) as $I_s^i$ and $I_{s}^{'i}$; random augment artist's name $t_s^i$ and $t_s^{'i}$.
    \State compute stylized output $I_{cs}$ and $I'_{cs}$;
    \EndFor
    \For{All $i\in \{1,...,N\}\} and \{j\in \{1,...,N\}$}
    \State compute $S(\nu_i^I, \nu_j^{'I})/\tau$ and $S(\nu_i^I, \nu_j^{'t})/\tau$;
    \EndFor
    \State compute contrastive similarity loss as Equation~\ref{eq:simclr}, and also other five losses for the final loss;
\EndFor
\State \Return \net 
\label{procedure}
\label{code}
\end{algorithmic}
\label{al:trainalgorithm1}
\end{algorithm}

\noindent We use VGG-19~\cite{VGG} to extract $W_s$ features (\textit{relu1\_2}, \textit{relu2\_2}, \textit{relu3\_4}, \textit{relu4\_1}) to compute the style loss, where $\mu$ and $\sigma$ are the mean and variance operators. We also extract $W_c$ features (\textit{relu2\_2}, \textit{relu3\_4}) to compute the content loss. Furthermore, identity loss $L_{id}$ is also used to preserve more content information. The process is to take the target image as both content and style input to extract the stylized feature $F_{cs}$, then compute the $L_1$-norm distance to the content feature $F_c$ as $L_{id}=\Vert F_{cs} - F_c\Vert_{1}^{1}$.

\noindent $\bullet$  \textbf{Total loss.} We train the network using all pre-defined losses. Hence, we define the final loss as $L=\lambda_{clip}L_{clip}+\lambda_{clip_f}L_{clip_f}+\lambda_{sim}L_{sim}+\lambda_{sty}L_{sty}+\lambda_{con}L_{con}+\lambda_{id}L_{id}$, where $\lambda_{clip},\lambda_{clip_f},\lambda_{sim},\lambda_{sty},\lambda_{con}$ and $\lambda_{id}$ are coefficients to balance these loss components. The whole training procedure is summarized in Algorithm~\ref{al:trainalgorithm1}.
\section{Experiments}






\subsection{Implementing Details}
\noindent $\bullet$  \textbf{Training strategy.} To achieve better visual quality, we use the VGG-19~\cite{VGG} pre-trained on the ImageNet dataset as the Image Encoder and fix its parameters. Then, we train \net in two stages: (1) We built an auto-encoder structure with the Image Encoder and Image Decoder (i.e., the orange box in Figure~\ref{fig:network}), and train the Image Decoder for image reconstruction. (2) We add the Positional mapper and Polynomial attention module to train the whole network for style transfer, using pre-trained Image Encoder and Decoder.

\noindent $\bullet$ \textbf{Datasets.} We use the images from MS-COCO~\cite{COCO} (about 118k images) for the image reconstruction task in the first training stage. In the second training stage, we train \net with MS-COCO~\cite{COCO} as our content image set and WikiArt~\cite{WikiArt} (about 81k images) as the style image set. In the training phase, we load the images with the size of $512\times512$ and randomly crop as training patches of size $256\times256$. As data augmentation, we randomly flip the content and style images. For inference, our \net can handle the images with any resolutions. In this Section, we use the images with $512\times512$ resolution for fair comparison. We also use WikiArt\_subset~\cite{ast}, a subset from WikiArt, to compare artist-aware style transfer. It contains 13 artists and each artist has 40-200 paintings.

\noindent $\bullet$ \textbf{Parameter setting.} We train  \net using Adam optimizer with the learning rate of $1\times10^{-4}$. The batch size is set to 30 and \net is trained for 100k iterations (about 8 hours) on a PC with two NVIDIA RTX3090 GPUs using PyTorch deep learning platform. The weighting factors in the total loss are defined as: $\lambda_{clip}=100, \lambda_{clip_f}=50, \lambda_{sim}=10, \lambda_{sty}=5, \lambda_{con}=1, \lambda_{id}=2$.

\noindent $\bullet$  \textbf{Metrics and evaluation.} Following VAEST~\cite{vaest} and LT~\cite{LT}, we estimate the content loss and style loss (Equation~\ref{eq:con_sty}) for VGG score comparison. We also conduct CLIP score estimation, CLIP loss (text/image-image cosine similarity)~\cite{language_2} to measure the semantic similarity between the target texts/images and stylized images. For artist-awareness, we follow AST~\cite{ast} to compute the style transfer deception rate, which is \textit{calculated as the fraction of generated images which are classified by the VGG-16 network as the artworks of an artist for which the stylization was produced.} Higher value means closer to the artist style.

\subsection{Comparison with the Arbitrary Style Transfer Methods}
Benefited from CLIP, the style clue of our \net can be either image or text. The arbitrary style transfer methods~\cite{wct, adain, sanet, artflow, adaattn, dstn, vaest} usually use style images to provide guidance for style transfer. We collect 20 content images and 20 style images to form 400 image pairs for image-driven arbitrary style transfer in this section. To demonstrate the effectiveness of \net on arbitrary style transfer, we use reference images to provide style clue for all methods.  We compared \net to seven state-of-the-art arbitrary style transfer methods, including WCT~\cite{wct}, AdaIn~\cite{adain}, SANet~\cite{sanet}, ArtFlow~\cite{artflow}, AdaAttN~\cite{adaattn}, DSTN~\cite{dstn}, and ST-VAE~\cite{vaest}.  

\begin{figure*}[t]
	\begin{center}
		\centerline{\includegraphics[width=2\columnwidth]{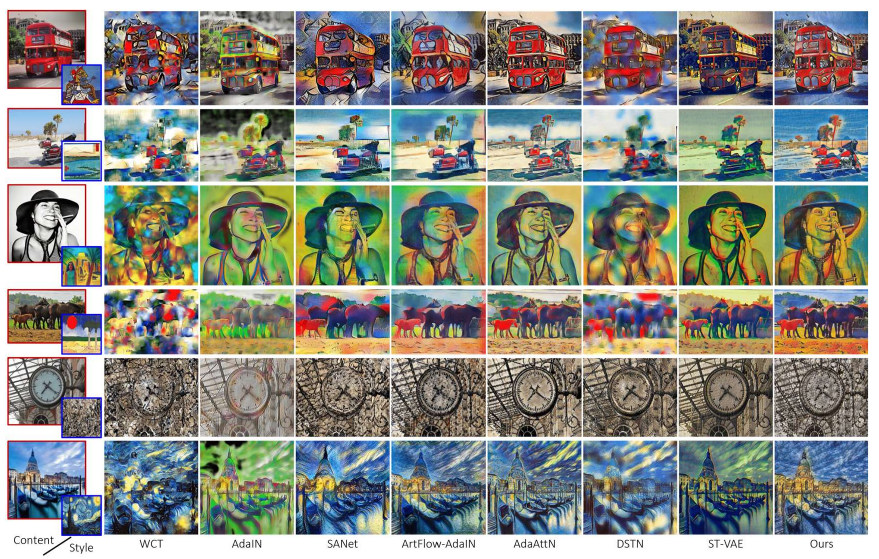}}
		\caption{\small
		\textbf{Comparison among image-driven arbitrary style transfer methods.} Our results have a good balance between the contents and styles with clear semantic information from the content image and painting preference from the style image. Please zoom in for better view.
		}
		\label{fig:SOTAs}
	\end{center}
\end{figure*}

\noindent \textbf{Quantitative and Qualitative Comparisons.}
Same as the VAEST~\cite{vaest} and LT~\cite{LT}, we use the content and style losses of Equation~\ref{eq:con_sty} to estimate the style and content performance. Let us consider them as the VGG score, since they are computed from the pre-trained VGG network. Lower content and style losses mean better performance ($\downarrow$ in the table). Feature extracted by the CLIP model has high-semantic discriminate power that can be used for similarity measurement. Therefore, we also compute CLIP scores. First we extract the feature embedding from the content, style, and transfer images by CLIP, then we compute content $s_{cont}$ and style $s_{style}$ scores, as defined by Equations~\ref{eqn:clip_cont} and \ref{eqn:clip_sytle}, for which the higher score means better performance ($\uparrow$ in the table).  Given that the similarity to the content and style reference images are at odds with each other, there is always a trade-off between them. $F1$ score is widely used in the binary classification field to measure the harmonic mean of the precision and recall~\cite{f1}. Inspired by it, we measure the harmonic mean (denoted it as $F1$ score) of the content and style similarity as defined in Equation~\ref{eqn:clip_f1}. Subjectively, we make visual comparisons among different approaches, as shown in Figure~\ref{fig:SOTAs}.

\begin{small}
\begin{equation} \tag*{(8)}
\label{eqn:clip_cont}
s_{cont}\left( I_c,I_{cs} \right) =\frac{E_{I}^{I_c}\cdot E_{I}^{I_{cs}}}{\left\| E_{I}^{I_c} \right\| \times \left\| E_{I}^{I_{cs}} \right\|}
\end{equation}

\vskip -0.05in
\begin{equation} \tag*{(9)}
\label{eqn:clip_sytle}
s_{style}\left( I_s,I_{cs} \right) =\frac{E_{I}^{I_s}\cdot E_{I}^{I_{cs}}}{\left\| E_{I}^{I_s} \right\| \times \left\| E_{I}^{I_{cs}} \right\|}
\end{equation}
\vskip -0.1in
\begin{equation} \tag*{(10)}
\label{eqn:clip_f1}
F1\left( I_c,I_s,I_{cs} \right) =2\times \frac{s_{style}\left( I_s,I_{cs} \right) \times s_{cont}\left( I_s,I_{cs} \right)}{s_{style}\left( I_s,I_{cs} \right) +s_{cont}\left( I_s,I_{cs} \right)}
\end{equation}

\end{small}

\begin{figure}[t]
	\begin{center}
		\centerline{\includegraphics[width=1\columnwidth]{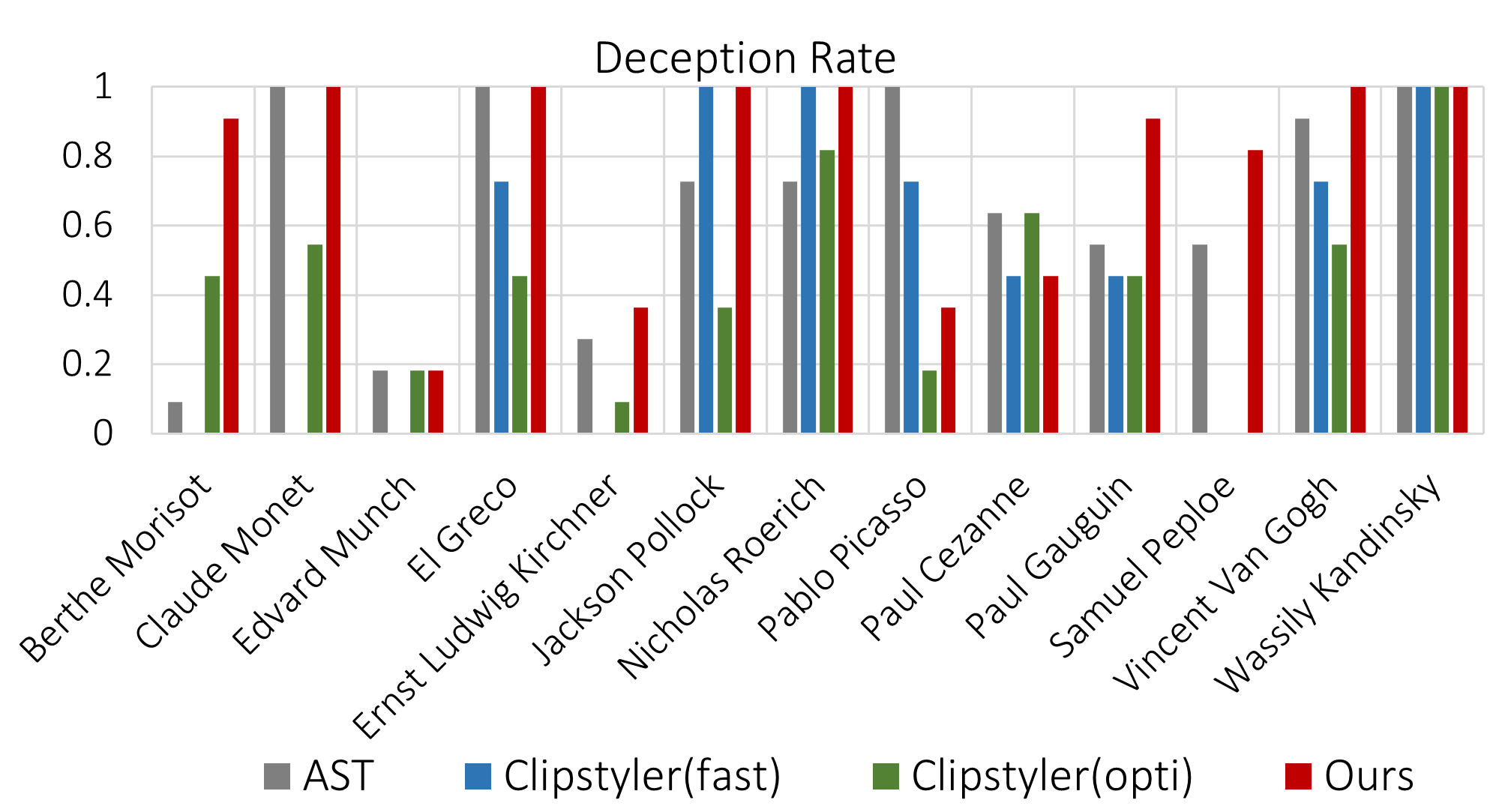}}
		\caption{\small
		\textbf{Deception rate for different artists.} We show the deception rate on each artist from WikiArt\_subset. Our method achieves the best performance on all artists. 
		}
		\label{fig:deception_artist}
	\end{center}
\end{figure}

\begin{table}[t]
\setlength\tabcolsep{4.2pt}
\centering
\caption{Comparison with the SOTA arbitrary style transfer methods. ({\color[HTML]{C00000}  {Red}}: best, {\color[HTML]{008000}  {Green}}: the $2^{nd}$ best, and {\color[HTML]{0000FF}  {Blue}}: the $3^{rd}$ best)} 
\label{tab:SOTA} 
\vskip -0.1in
\begin{tabular}{cccccc}
\toprule
 &
   \multicolumn{2}{c}{\textit{VGG Score}}
   &
  \multicolumn{3}{c}{\textit{CLIP Similarity Scores}} \\
\multirow{-2}{*}{ {Method}} &
  {Style$\downarrow$} &
  {Content$\downarrow$} &
   {Content$\uparrow$} &
   {Style$\uparrow$} &
   {F1$\uparrow$} \\ \midrule
WCT\cite{wct} &
  {\color[HTML]{C00000}  {74.11}} &
  197.26 &
  0.487 &
  {\color[HTML]{C00000}  {0.729}} &
  0.584 \\
AdaIN\cite{adain} &
  167.53 &
  {\color[HTML]{0000FF}  {123.02}} &
  {\color[HTML]{008000}  {0.642}} &
  0.599 &
  0.620 \\
SANet\cite{sanet} &
  104.41 &
  153.65 &
  0.541 &
  {\color[HTML]{008000}  {0.700}} &
  0.611 \\
ArtFlow-AdaIN\cite{artflow}&
  149.98 &
  138.59 &
  0.578 &
  0.647 &
  0.611 \\
AdaAttN\cite{adaattn} &
  {\color[HTML]{008000}  {83.45}} &
  {\color[HTML]{008000}  {119.28}} &
  {\color[HTML]{C00000}  {0.658}} &
  0.604 &
  {\color[HTML]{0000FF}  {0.630}} \\
DSTN\cite{dstn} &
  122.34 &
  130.11 &
  0.616 &
  0.663 &
  {\color[HTML]{008000}  {0.638}} \\
ST-VAE\cite{vaest} &
  108.73 &
  139.23 &
  {\color[HTML]{0000FF}  {0.635}} &
  0.581 &
  0.606 \\ \midrule
 \textbf{Ours} & 
  {\color[HTML]{0000FF}  \textbf{99.36}} &
  {\color[HTML]{C00000}  \textbf{82.72}} &
  \textbf{0.637} &
  {\color[HTML]{0000FF}  \textbf{0.791}} &
  {\color[HTML]{C00000}  \textbf{0.706}} \\ \bottomrule

\end{tabular}
\end{table}

Table~\ref{tab:SOTA} shows the results of the quantitative comparison, while Figure~\ref{fig:SOTAs} illustrates the quantitative comparison. WCT~\cite{wct} utilizes the whitening and coloring transforms to the content features to match the statistical distribution of the style features. We observe that WCT~\cite{wct} has the lowest style loss (74.11) and the highest style similarity score (0.729) to the style image. The best style scores suggest that WCT is good at handling global style synthesis. However, WCT lacks attention to details. The visualization in Figure~\ref{fig:SOTAs} clearly shows the results produced by WCT have many improper style patches, and the contents are distorted significantly (see 2nd, 3rd and 4th rows). Consequently, this leads to lower content performance with a content loss of 197.26 and a content similarity score of 0.487. AdaAttN~\cite{adaattn} has the best content similarity score of 0.658. Compared to WCT, the stylizations of AdaAttN have clearer contents as shown in Figure~\ref{fig:SOTAs}, but they are limited in expressing target major color changes and texture synthesis. For instance, the background of the bus (see 1st row) should be dominated by blue, and the texture of the clock is not clear (see 5th row). 
Instead, our \net successfully transfers the target style well, and preserve the salient content at the same time. 
As shown in Figure~\ref{fig:SOTAs}, our stylized results clearly show the contents like that in AdaAttN (see the bus in 1st row, the motorcycle in 2nd row), while the style stays on par with WCT's and SANet's. Besides, the second row of Figure~\ref{fig:SOTAs} shows that the result with \net sufficiently reflects the semantic of the content image, and expresses the painting preference of the style image (like a woodblock painting). Quantitatively, our approach achieves the best balanced performance with an average F1 score of 0.706, as shown in Table~\ref{tab:SOTA}.

\begin{figure*}[t]
	\begin{center}
		\centerline{\includegraphics[width=2\columnwidth]{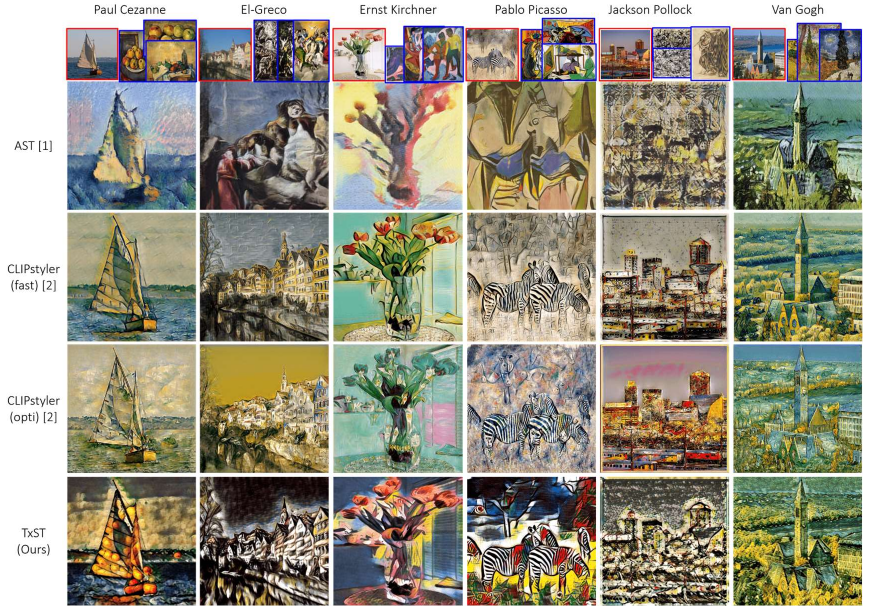}}
		\caption{\small
		\textbf{Comparison among text-driven artist-aware style transfer methods.} We show six examples (content images in red boxes) using three different methods: AST~\cite{ast}, CLIPstyler~\cite{language_2} and ours. We use names of painters as text prompts for style transfer. For reference, we also list some representative paintings in blue boxes for comparison. Styles of artists are very abstract and subjective. We highly recommend readers to check each artist online or from our supplementary files for better comparison.
		}
		\label{fig:artist_compare}
	\end{center}
\end{figure*}

\subsection{Artistic Style Transfer}
Artistic styles are more complex, abstract and diverse. It is difficult to have one reference image to represent the artist's signature style. One of our key claims is to learn artistic styles using specific artists' names. To show the ability of our proposed approach, we fine-tune \net on the WikiArt subset used in AST~\cite{ast}, which contains 13 artists of different styles. We also compare it to the recent text-driven style transfer CLIPstyler. For text-driven style transfer, we use the names of the 13 artists as style description (denoted as $T_s$). Then, following the design in \cite{language_2}, we train CLIPstyler with fast and optimized schemes, and denote them as CLIPstyler(fast) and CLIPstyler(opti). 

\begin{table}[t]
\linespread{1.2}
\caption{{\textbf{Comparison with different Artistic Style Transfer methods.}\\ ({\color[HTML]{C00000}  {Red}}: best, {\color[HTML]{0000FF}  {Blue}}: the $2^{nd}$ best).}
	}
	\label{tab:artists}
	\vskip -0.1in
\setlength\tabcolsep{7.2pt}
\centering
\begin{tabular}{ccccc}
\hline
                                  & \multicolumn{3}{c}{Clip Scores}                                                    & Deception           \\
\multirow{-2}{*}{Method} & Content$\uparrow$             & Style$\uparrow$               & F1$\uparrow$                 & Rate$\uparrow$                \\ \hline
AST\cite{ast}                               & 0.538                        & 0.269                        & 0.359                        & {\color[HTML]{0000FF} 0.664} \\
CLIPstyler(fast)\cite{language_2}                  & {\color[HTML]{C00000} 0.736} & 0.254                        & 0.378                        & 0.469                        \\
CLIPstyler(opti)\cite{language_2}                   & 0.624                        & {\color[HTML]{C00000} 0.326} & {\color[HTML]{C00000} 0.428} & 0.441                        \\
\textbf{Ours} &
  {\color[HTML]{0000FF} \textbf{0.678}} &
  {\color[HTML]{0000FF} \textbf{0.313}} &
  {\color[HTML]{0000FF} \textbf{0.418}} &
  {\color[HTML]{C00000} \textbf{0.769}} \\ \hline
\end{tabular}
\end{table}

\noindent \textbf{Quantitative Comparison.} 
Here, we measure the similarity to the content and artist in the CLIP feature space using the CLIP scores defined in Equations~\ref{eqn:clip_cont} and \ref{eqn:clip_text}, and compute the F1 score accordingly. For artist-awareness, we follow AST~\cite{ast} to compute the style transfer deception rate. Specifically, we use a trained VGG-16 model to classify the transferred images into different artists, where the classification accuracy is denoted as the deception rate. A higher deception rate means the transferred images 
have better artist stylization. 

\begin{small}
\begin{equation} \tag*{(11)}
\label{eqn:clip_text}
s_{style}\left( T_s,I_{cs} \right) =\frac{E_{I}^{T_s}\cdot E_{I}^{I_{cs}}}{\left\| E_{I}^{T_s} \right\| \times \left\| E_{I}^{I_{cs}} \right\|}
\end{equation}
where $T_s$ denotes the style description in the form of text.
\end{small}

Table~\ref{tab:artists} reports the results. We observe that CLIPstyler(fast) leads to the best content similarity score (0.736); however,
the transferred images have poor artistic style performance, with a deception rate of 0.469. The results of AST have the second-best deception rate of 0.664, but the similarity to the content image is of 0.538. 
CLIPstyler(opti) has the best CLIP scores, but achieves the worst deception rate. This is expected, since CLIPstyler(opti) requires dedicated training for each individual content and style dataset. It will optimize the CLIP scores as its loss function. However, it also introduces bizarre patterns (as shown in Figure~\ref{fig:teaser} and \ref{fig:artist_compare}) into the stylization. The results are not consistent to each specific style. 
Instead, our \net provides a good balance to the style and content, reaching the second-best performance on CLIP scores and the best performance on the deception rate, which demonstrates its efficiency and effectiveness. We also note that AST, CLIPstyler(opti) and CLIPstyler(fast) train different models for different artists; in cotrast, \net is used for \emph{arbitrary} style transfer, which is more flexible for applications. Figure~\ref{fig:deception_artist} also shows the class-specific accuracy of deception rate. We  clearly observe that using our \net (red bars) achieves dominant performance on all artists. 

\begin{figure}[t]
	\begin{center}
		\centerline{\includegraphics[width=1\columnwidth]{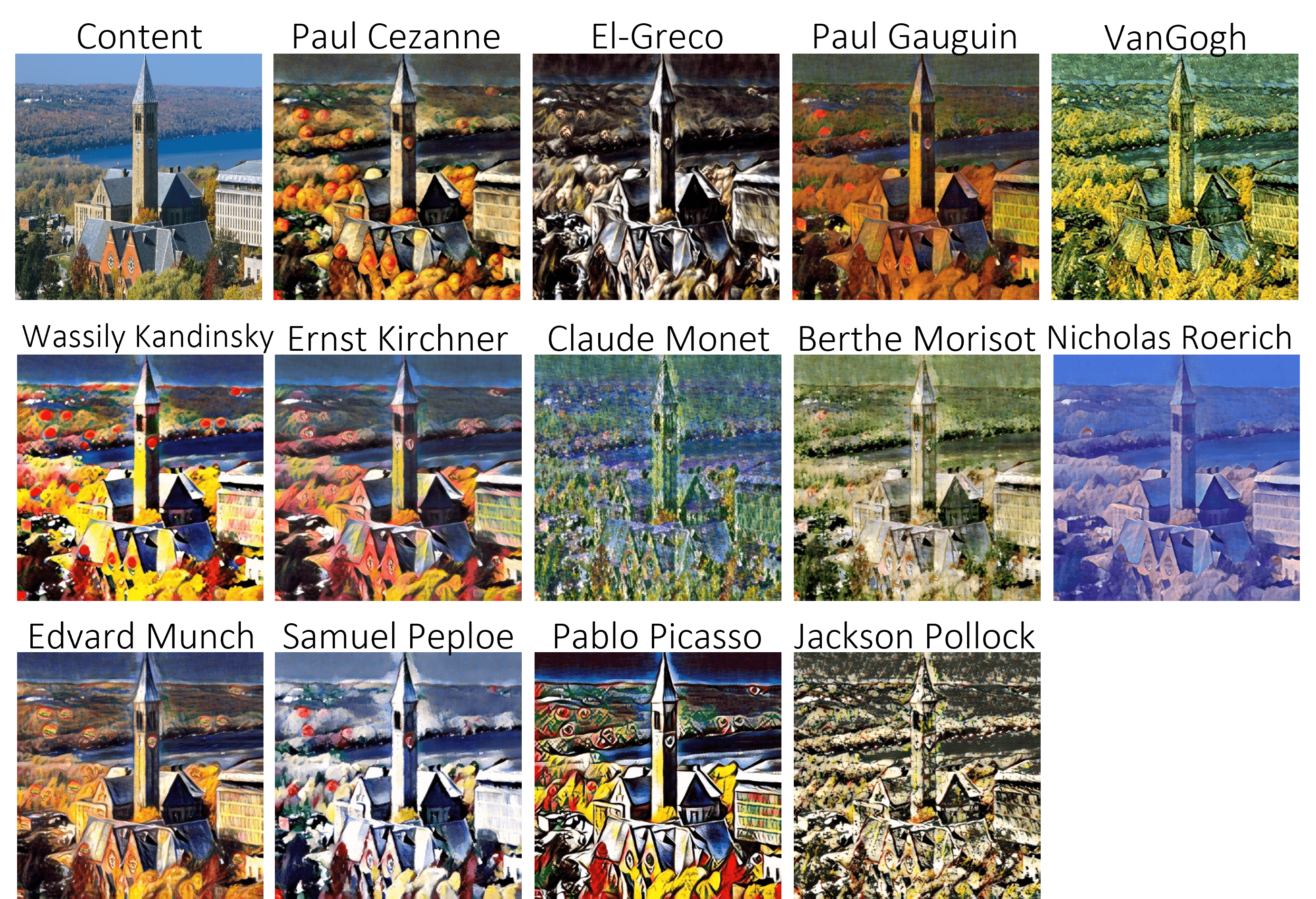}}
		\caption{\small
		\textbf{Visualization of 13 artistic style transfers.} We used the name of 13 artists in WikiArt\_subset as text input and used our proposed \net for style transfer.
		}
		\label{fig:artist_compare_2}
	\end{center}
\end{figure}

\begin{figure}[t]
	\begin{center}
		\centerline{\includegraphics[width=1\columnwidth]{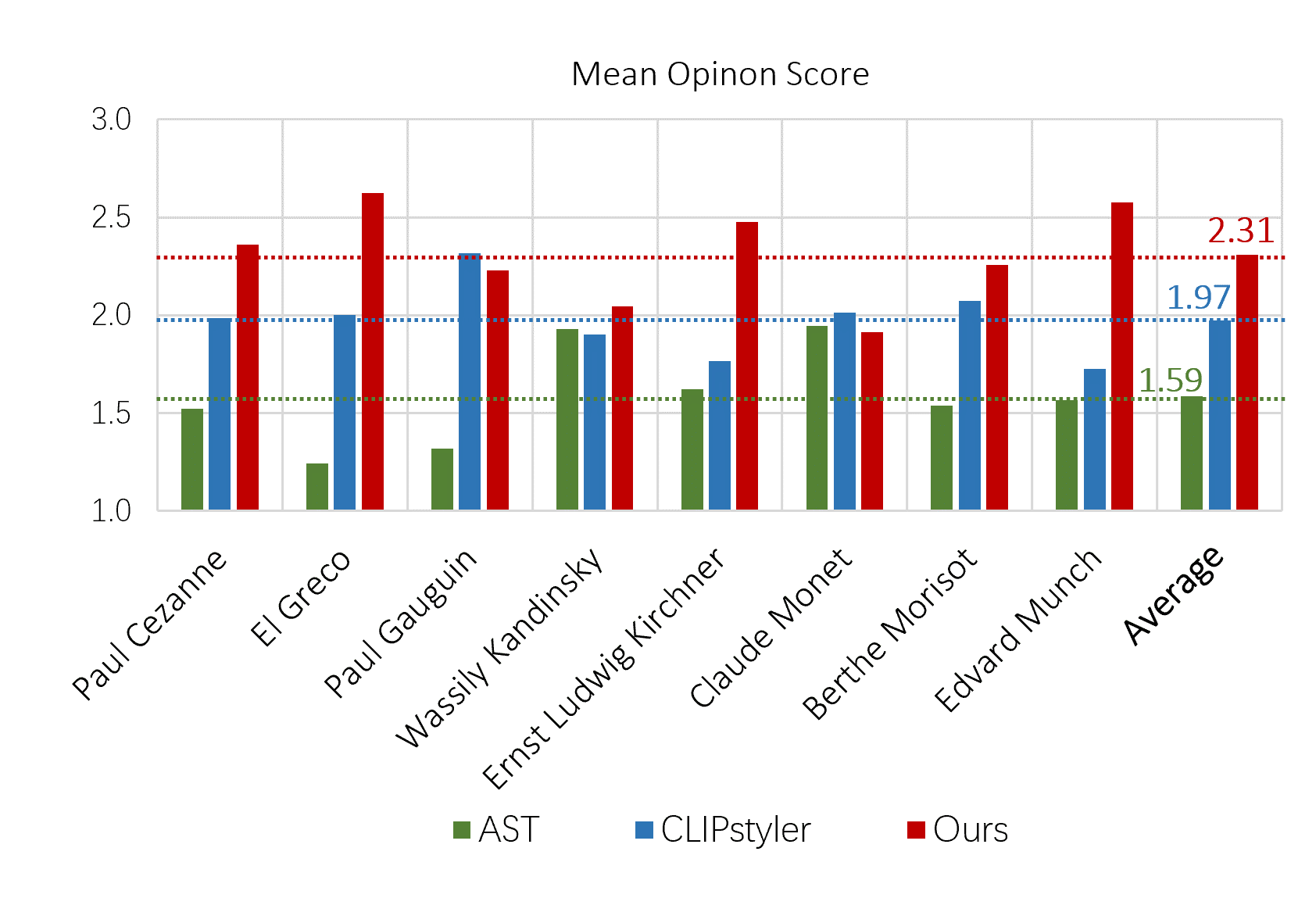}}
		\caption{\small
		\textbf{User study.}  We used the name of eight artists in WikiArt\_subset as text input for style transfer. We invited users to rank the results from AST~\cite{ast}, CLIPStyler(opti)~\cite{language_2}, and our proposed \net for artistic style transfer. The higher Mean Opinion Score (MOS), the more style similarity to the target artists.
		}
		\label{fig:usr_study}
	\end{center}
\end{figure}

\noindent \textbf{User study.}
Since artistic styles are very subjective, we also conduct a user study for subjective evaluation. We invite users from different backgrounds, like art, design, literature, and science, to ensure the study as fair as possible. The questionnaire compares the painting styles of eight artists from the WikiArt\_subset. For each artist, we collect the results from AST~\cite{ast}, CLIPstyler(opti)~\cite{language_2}, and the proposed TxST by using three different content images. In each question, the users were given three results from the same content image by using the three methods. Besides the artists' names, we provide users with three representative artworks from the artist as references. The users are requested to rank the style similarity to the artist, where 3 denotes the most similar, and 1 means the most different. We then average the scores from all users as the Mean Opinion Score (MOS). The results are shown in Figure~\ref{fig:usr_study}. We observe that \net achieves the highest average MOS of 2.31, which suggests that our results have the most similar painting style to the target artists \emph{at a perceptual level}.

\noindent \textbf{Qualitative Comparison.}
Figure~\ref{fig:artist_compare} shows the visual comparison of different methods for artist-aware style transfer. There are six content images in red boxes and six artists' names for style transfer. Note that AST is an optimization based approach that learns a dedicated model for every artist. It does not require any name as text input. For CLIPstyler and \net (ours), we use artists' names as textual style to guide stylization. For reference, we also show three representative paintings in blue boxes for six artists. Our findings are summarized as: First, our results show comparable or even better content preservation ability as compared to AST and CLIPstyler. For instance, like CLIPstyler, \net can restore structural information, like buildings. It can also preserve the fine textures, like the strides on zebras. Second, \net can faithfully mimic the signature styles from specific artists. For example, our approach can ``copy'' the fruit patterns from \textit{Paule Cezanne}, especially the color tone and temperature, while other approaches fail. Moreover, \net can transfer painting curvatures, like the distorted curves in \textit{El-Greco} and \textit{Van Gogh}, and can learn the color mosaic from \textit{Ernst Kirchner}. On the other hand, AST and CLIPstyler (both fast and optimal versions) do not reveal the key features of artists, like zebras in \textit{Pablo Picasso} style and buildings in \textit{Jackson Pollock} style. Third, compared to CLIPstyler, \net can maximize the visual differences among different artistic styles (this is further discussed in Section~\ref{sub:ablation}). Finally, \net is a universal style transfer without requiring neither dedicated training for each style (as AST and CLIPstyler(fast)) nor extra online training efforts (as CLIPstyler(opti)).


\subsection{Ablation Study}
\label{sub:ablation}
To confirm the effectiveness of each component of \net, we conduct several ablation studies.

\noindent $\bullet$  \textbf{Loss terms.} First, as introduced in Section~\ref{sub:losses}, we have six different loss terms for training. Content loss, style loss and identity loss, which are widely used in style transfer to encourage stylization. The key terms are Contrastive similarity loss ($L_{sim}$), Directional CLIP loss ($L_{clip}$) and CLIP feature loss ($L_{clip_f}$). To evaluate their effects, we use the 13 artists' names from WikiArt\_subset as input for text based style transfer. Our metrics include content loss, VGG feature loss between content and stylized images (as $L_{con}$ in Equation~\ref{eq:con_sty}), text CLIP loss (Equation~\ref{eqn:clip_text}) and content CLIP loss (Equation~\ref{eqn:clip_cont}). 


\begin{table}[t]
	\caption{{\textbf{Comparison on loss terms.} for text based style transfer in images from COCO~\cite{COCO} using artists from WikiArt\_subset.}
	}
	\label{tab:abla_loss}
	\vskip -0.1in
	\vspace{-2mm}
	\begin{center}
		\begin{small}
			\scalebox{0.85}{
\begin{tabular}{c|cccc}
\toprule
\multirow{2}{*}{Metric}   & \multirow{2}{*}{Content$\downarrow$} & \multicolumn{3}{c}{CLIP Similarity Score} \\
& & Text$\uparrow$ & Cont$\uparrow$ & F1$\uparrow$ \\ \midrule
\begin{tabular}[c]{@{}c@{}} $L_{con}+L_{sty}+L_{id}$($baseline$)\end{tabular} & 86.82        & 0.207     & 0.702  & 0.320  \\ \midrule
$baseline+L_{clip}$                                                         & 86.89        & 0.254     & 0.575   & 0.352 \\
$baseline+L_{sim}$                                                         & 86.97        & 0.261     & 0.530   &  0.350\\
$baseline+L_{clip_f}$                                                        & 86.78        & 0.216     & 0.706  &  0.331 \\
$baseline+L_{sim}+L_{clip}$                                                     & 87.19        & 0.295     & 0.525  & 0.378  \\
$baseline+L_{clip}+L_{clip_f}$                                                    & 86.86        & 0.238     & 0.660 &  0.350  \\
$baseline+L_{sim}+L_{clip_f}$                                                     & 86.79        & 0.244     & 0.645  &  0.354 \\
\cellcolor{mistyrose}{$baseline+L_{clip}+L_{clip_f}+L_{sim}$}                                                & \cellcolor{mistyrose}{86.90}        & \cellcolor{mistyrose}{0.313}     & \cellcolor{mistyrose}{0.677}   & 
\cellcolor{mistyrose}{0.428} \\ \bottomrule
\end{tabular}
			}
		\end{small}
	\end{center}
	\vskip -0.2in
\end{table}

Table~\ref{tab:abla_loss} shows the results when training with different losses. 
\textit{Baseline} is the model that uses content ($L_{con}$), style ($L_{sty}$) and identity losses ($L_{id}$) and compare it with different loss combinations. We observe that using
both Contrastive similarity loss ($L_{sim}$) (3rd row) and Directional CLIP loss (2nd row) reduces the text CLIP loss, which indicates that they can guide the stylization close to the target text description. Meanwhile, using CLIP feature loss ($L_{clip_f}$) (4th row) reduces the content loss and content CLIP loss approximately by 0.05 and 0.1, respectively, but it also increases the text CLIP loss. The last row is our \net{full \net} model that combines all loss terms. 
Obtaining the best results is a balance between content and text styles.

\noindent $\bullet$  \textbf{Polynomial attention module.} \net introduces the polynomial attention module to explore high-order correlations between content and style features, hence it generates images with styles closer to the style references. As described in equation~\ref{eq:Equation3}, we can compute the cross correlation from the  $1^{st}$ to $n^{th}$ order features. To quantitatively decide the optimal polynomial setting, we use 20 style images as reference and 20 content images as target for image based style transfer. To demonstrate its effectiveness, we compute different metrics and report their results in Table~\ref{tab:abla_attn}.

Specifically, in Table~\ref{tab:abla_attn} we compare both VGG based losses and CLIP based losses. Compared to the $1^{st}$ order model, using other polynomial models higher than $1^{st}$ order improves both content and style performance, except the $5^{th}$ order (last row), which indicates that using higher order polynomial can indeed improve the overall performance, but it may also drop after reaching the optimal point. This is expected, since higher-order models mean more detailed regressive modelling, but they can also overfit the correlation between content and style images. Moreover, we observe  that using higher order polynomials reduces the style loss, but it also increases the content loss. This is the trade-off between content and style, as discussed in~\cite{adaattn,LT,art-net,artflow}. For practical applications, we use the $2^{ed}$ polynomial model for its good balance between content and style (higher F1 score compared to others).

\begin{table}[t]
\setlength\tabcolsep{4.6pt}
\centering
\caption{\textbf{Comparison on using different orders of Polynomial attention for style transfer.} Both VGG and CLIP based scores are used for evaluation.} 
\label{tab:abla_attn} 
\begin{tabular}{cccccc}
\toprule
 &
   \multicolumn{2}{c}{\textit{VGG Score}}
   &
  \multicolumn{3}{c}{\textit{CLIP Similarity Score}} \\
\multirow{-2}{*}{ {Order of Attention}} &
  {Style$\downarrow$} &
  {Content$\downarrow$} &
   {Content$\uparrow$} &
   {Style$\uparrow$} &
   {F1$\uparrow$} \\ \midrule
1st order &
  100.28 &
  91.05 &
  0.612 &
  0.683 &
  0.648 \\
\cellcolor{mistyrose}{2nd order} &
  \cellcolor{mistyrose}{99.36} &
  \cellcolor{mistyrose}{82.72} &
  \cellcolor{mistyrose}{0.37} &
  \cellcolor{mistyrose}{0.791} &
  \cellcolor{mistyrose}{0.706} \\
3rd order &
  100.41 &
  84.40 &
  0.620 &
  0.787 &
  0.693 \\
4th order &
  101.97 &
  84.74 &
  0.618 &
  0.797 &
  0.696 \\
5th order &
  103.00 &
  87.99 &
  0.616 &
  0.717 &
  0.663 \\ \bottomrule

\end{tabular}
\end{table}

\begin{figure}[b]
	\begin{center}
		\centerline{\includegraphics[width=1\columnwidth,keepaspectratio]{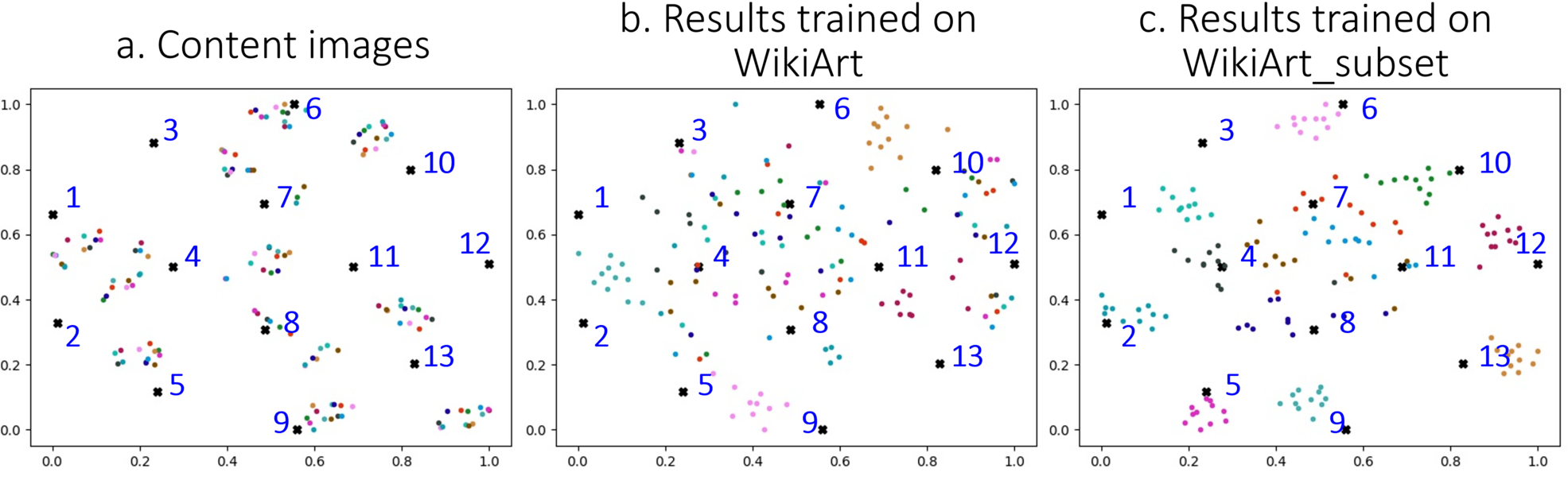}}
		\caption{\small
		\textbf{t-SNE visualization on WikiArt style transfer.} Symbols ``x'' are the style texts, dots in different colors are the stylized images. 13 X represents 13 artists from WikiArt\_subset. Figure (a) is the distribution of the content images, (b) is the distribution of the styled images obtained by the model trained on whole WikiArt, (c) is the distribution of the styled images obtained by the model trained on WikiArt\_subset.
		}
		\label{fig:tsne}
	\end{center}
\end{figure}

\begin{figure*}[ht!]
	\begin{center}
		\centerline{\includegraphics[width=1\textwidth,height=0.8\textheight]{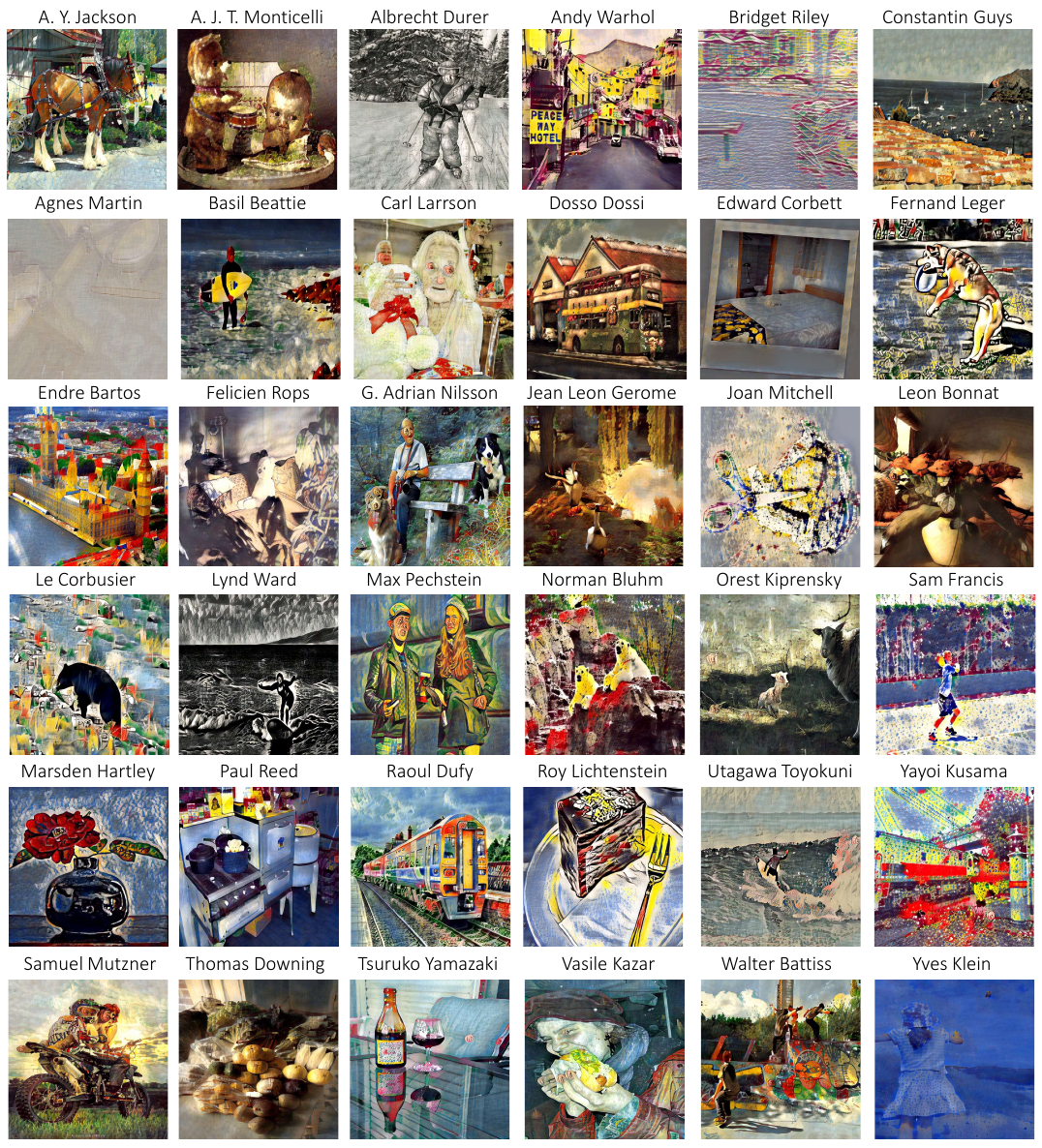}}
		\caption{\small
		\textbf{Style comparison of different artists from WikiArt dataset.} We selected 18 artists' names from WikiArt as style input and 18 content images for style transfer. 
		}
		\label{fig:wiki_all}
	\end{center}
\end{figure*}

\noindent $\bullet$  \textbf{Analysis on the diversity of artist-aware style transfer.} 
All quantitative and qualitative comparisons so far demonstrate that \net generates stylized images with better visual quality than the state of the art. 
Here, however, we answer the question whether `styles of different artists are well separated to each other'?.

To analyze the style diversity of \net, we use t-SNE~\cite{t-sne} to process both content and stylized images and check the data scatters or the distribution. The basic process is: first, we use CLIP to extract feature codes for content images ($E_I(I_{c})$), style texts ($E_T(t)$) and stylized images ($E_I(I_{cs})$). Second, we use t-SNE to project the codes into the 2-D space and draw them on an x-y plane for visualization.

In Figure~\ref{fig:tsne}, we show both style texts (symbol ``X'') and images (symbols ``.''). Dots in different colors represent different images. We make three observations: (1) the style texts are well separated in the 2-D space, which indicates that CLIP has the ability to distinguish different style descriptions. (2) By comparing Figure~\ref{fig:tsne}~(a) to Figure~\ref{fig:tsne}~(b) and (c), we observe that the stylized images are well scattered around the space. This indicates that the stylization is successful for images, as even two different stylization results applied to the same initial image can be pushed away. (3) By comparing Figure~\ref{fig:tsne}~(b) to  Figure~\ref{fig:tsne}~(c), we observe that using the model trained on WikiArt already distinguishes different styles; however, using the model trained on WikiArt\_subset can push the images even further from each other. 


Furthermore, Figure~\ref{fig:wiki_all} displays more examples of artist-aware style transfer, thus corroborating style diversity. We observe that \net successfully grasps the key features of different artists and transfers them to the content images. The differences among different artists are significant. For example, artists \textit{A. J. T. Monticelli}, \textit{Constantin Guys}, \textit{Jean Leon Gerome} and \textit{Leon Bonnat} are from the late 18th to the early 19th century, when the main trend was religious painting and the painters used natural lightning and shadows to describe the sense of space. Painters like \textit{Andy Warhol}, \textit{Bridget Riley}, \textit{Joan Mitchell} are contemporary artists. They are bold and experimental. They usually use more colorful pigments, wild and rough curves and lines to paint their feelings. We also note some distinguish oriental styles from \textit{Utagawa Toyokuni}, who is one of the painters from the Ukiyo-e moment in Japan. For other painters, we encourage the readers to check more information from the WikiArt website \url{https://www.wikiart.org/}.

\subsection{Interactive Text-driven style transfer}
\net leads to flexible universal text-driven style transfer. The texts can be general descriptions like texture patterns, color distributions and objects. 
Moreover, it can achieve multiple style transfer by using texts; here, we explore this.

\noindent $\bullet$ \textbf{General style transfer.}
\net can take arbitrary texts as input for style transfer. The texts can describe objects or colors, like \textit{red apples} and \textit{colorful rainbow}. To achieve this, we slightly fine-tune \net using smaller weights on the CLIP loss (Equation~\ref{eq:clip}) so that the model is more sensitive to the changes of texts. Note that \net is still trained on the WikiArt dataset without using any extra labeled texts or images, hence the model is general to some extent. For any specific text, it still requires more dedicated training. 

\begin{figure}[t]
	\begin{center}
		\centerline{\includegraphics[width=1\columnwidth]{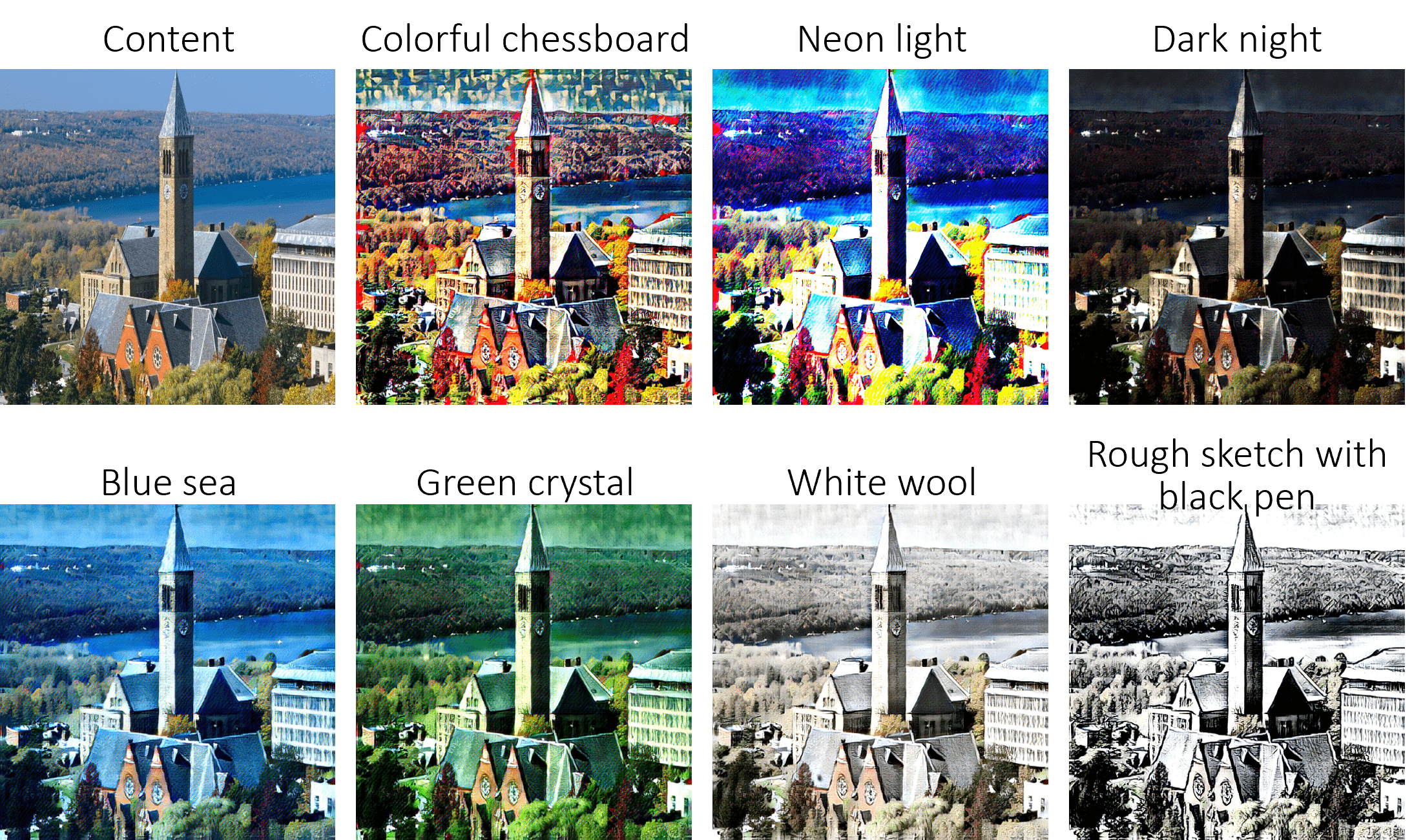}}
		\caption{\small
		\textbf{Visualization of arbitrary text-driven style transfer.} We used some random texts, like objects and colors, as style input for style transfer.
		}
		\label{fig:general_st}
	\end{center}
\end{figure}

\begin{figure}[t]
	\begin{center}
		\centerline{\includegraphics[width=1\columnwidth]{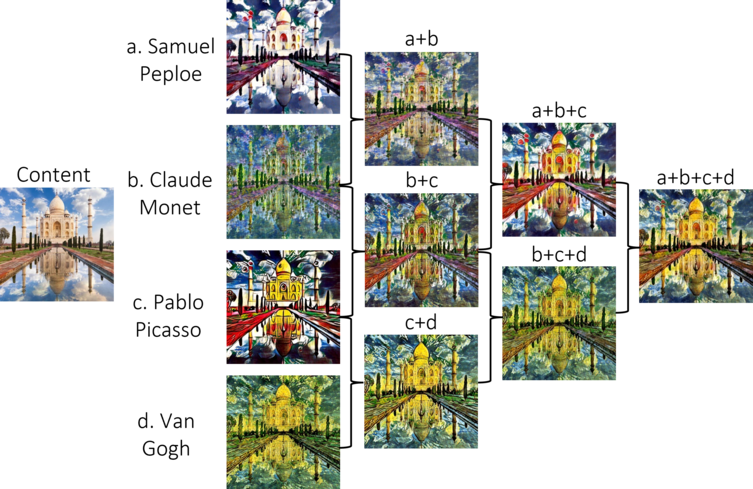}}
		\caption{\small
		\textbf{Visualization of multiple style transfer.} We use some random artists' names from WikiArt\_subset as input for multiple style transfer.
		}
		\label{fig:multiple_st}
	\end{center}
\end{figure}

For this, we test the content image using texts to describe different styles, like colors (\textit{blue sea, green crystal}), textures (\textit{white wool, colorful chessboard}) and visual appearance (\textit{rough sketch with black pen, neon light, dark night}). Figure~\ref{fig:general_st} illustrates the results. We observe that \net successfully transfers the target styles to the content images.

\noindent $\bullet$ \textbf{Multiple style transfer.}
\net can also achieve multiple style transfer. Unlike previous works~\cite{vaest,adain}, it does not require multiple reference images for style fusion; instead, we can give a description that combines multiple targets together as text input. 
In Figure~\ref{fig:multiple_st}, we use four artists' names as input for style fusion: \textit{Samuel Peploe, Claude Monet, Pablo Picasso, Van Gogh}. We first show the individual style transfer, then we combine different artists' names, like \textit{Samuel Peploe and Claude Monet} (a+b), \textit{Claude Monet and Pablo Picasso and Van Gogh} (b+c+d), we can see that the proposed \net can combine different art styles significantly. 

\section{Conclusion}


In this paper, we proposed a text-driven approach for artist-aware style transfer, coined \net. Unlike most style transfer methods, \net can use either images or texts as style references for the desired stylization. The polynomial attention module and CLIP-space-based contrastive similarity training of \net enable exploring the co-linearity between texts and images without requiring costly data collection and annotation. We conducted comprehensive experiments on both general and artist-specific style transfer, using either images or text as style references. Extensive results show that \net achieves perceptually pleasing arbitrary stylization, revealing its ability to extract critical representations from the CLIP space and produce aesthetics close to the artists' works. \net also points a new direction to text-driven style transfer. Future work includes combining images, texts, and other cues to deliver a more flexible user-guided style transfer.

\section*{Acknowledgements}
This work was supported by ANR-22-CE23-0007, Hi!Paris collaborative project and DIM RFSI 2021.

\ifCLASSOPTIONcaptionsoff
  \newpage
\fi



%


\bibliographystyle{IEEEtran}
\bibliography{main_PAMI}
%

\vspace{-8mm}
\begin{IEEEbiography}[{\includegraphics[width=1in,height=1.25in,clip,keepaspectratio]{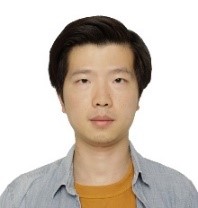}}]{Zhi-Song Liu}
received his M.Sc. (2015) and Ph.D. (2020) degrees from The Hong Kong Polytechnic University under the supervision of Prof. Wan-Chi Siu and Dr. Yui-Lam Chan. Dr. Liu is now a Post-Doctoral Fellow with Caritas Institute of Higher Education and was also with Ecole Polytechnique in 2020. His research interests include image and video signal processing, deep learning.
\end{IEEEbiography}
\vspace{-8mm}
\begin{IEEEbiography}[{\includegraphics[width=1in,height=1.25in,clip,keepaspectratio]{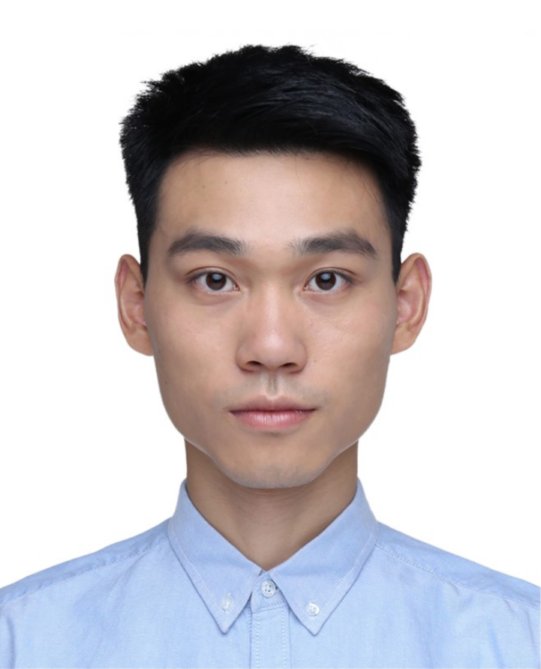}}]{Li-Wen Wang}
received his Bachelor of Engineering degree, in 2016, from Shandong University. Subsequently, he obtained his Master of Science (MSc) degree with distinction from The Hong Kong Polytechnic University, where he is now a final year Ph.D. candidate under the supervision of Professor Wan-Chi Siu and Dr. Daniel P. K. Lun. His research interests include deep learning, image and video processing, object detection and tracking.
\end{IEEEbiography}
\vspace{-8mm}
\begin{IEEEbiography}[{\includegraphics[width=1in,height=1.25in,clip,keepaspectratio]{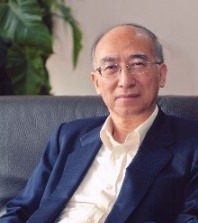}}]{Prof. Wan-Chi Siu}
is Emeritus Professor (formerly Chair Professor, HoD(EIE) and Dean of Engineering Faculty) of Hong Kong Polytechnic University and Research Professor of Caritas Institute of Higher Education in Hong Kong. He is Life-Fellow of IEEE and was a PhD graduate (1984) of the Imperial College London, Vice President, Chair of Conference Board and core member of Board of Governors of the IEEE SP Society (2012-2014) and Past President of Asia Pacific Signal and Information Processing Association (2017-2018), and has been Guest Editor/Subject Editor/AE for IEEE Transactions on CAS, IP, CSVT, and Electronics Letters. Prof. Siu has been Keynote Speaker and Invited Speaker of many conferences, published over 500 research papers in DSP, transforms, fast algorithms, machine learning, deep learning, super-resolution imaging, 2D/3D video coding, object recognition and tracking, and organized IEEE society-sponsored flagship conferences as TPC Chair (ISCAS1997) and General Chair (ICASSP2003 and ICIP2010). Recently, he has been a member of the IEEE Educational Activities Board, the IEEE Fourier Award for Signal Processing Committee (2017-2020), the Hong Kong RGC Engineering Panel Member-JRS (2020-2022) and some other IEEE technical committees.
\end{IEEEbiography}
\vspace{-8mm}
\begin{IEEEbiography}[{\includegraphics[width=1in,height=1.25in,clip,keepaspectratio]{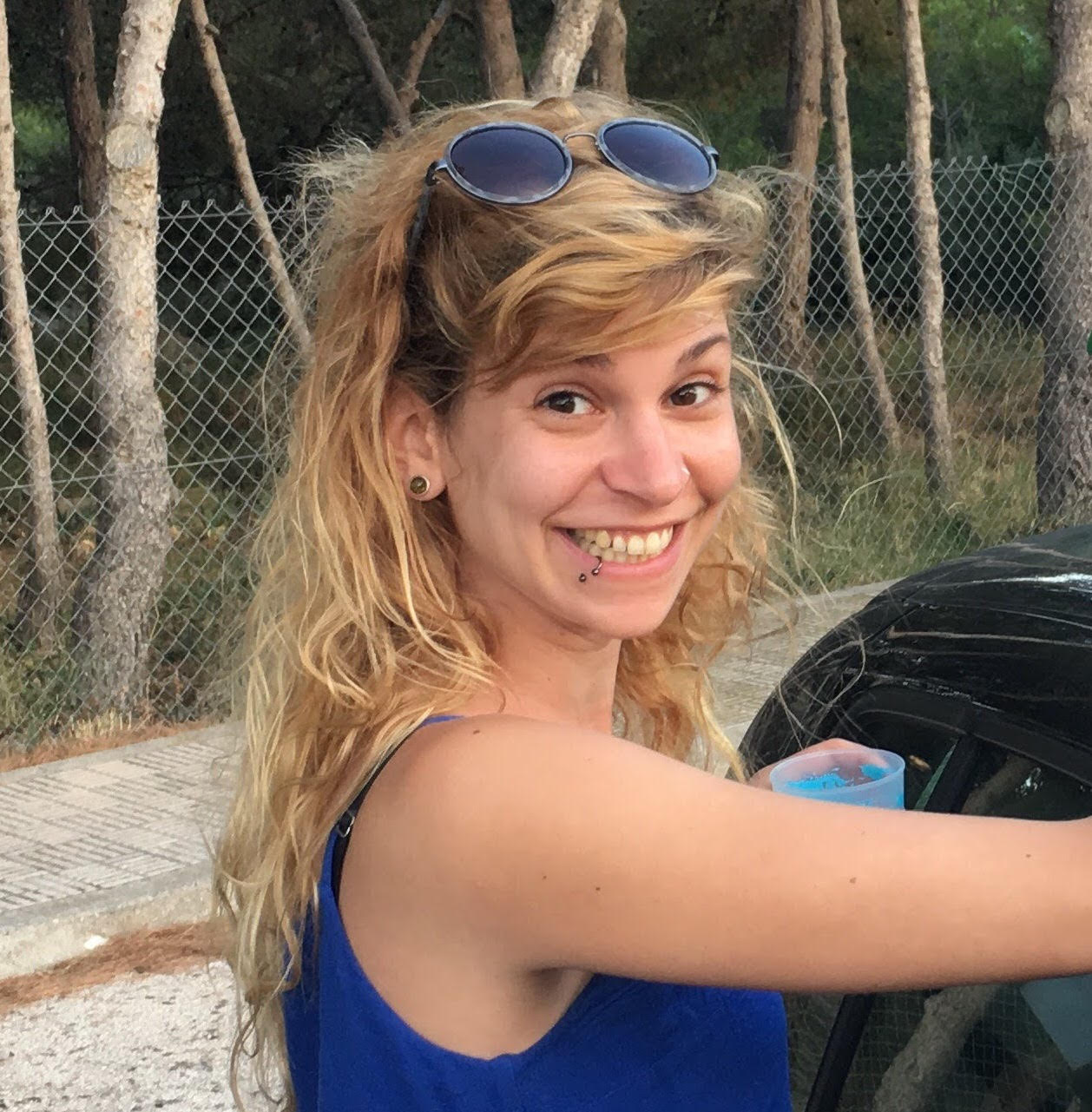}}]{Vicky Kalogeiton}
is an Assistant Professor at the GeoViC team of LIX, Ecole Polytechnique, IP Paris. Previously, she was a research fellow at Visual Geometry Group, University of Oxford, where she was working with Andrew Zisserman. 
She received her PhD from the University of Edinburgh and INRIA Grenoble in 2017. Her research interests are computer vision and, specifically, human-centric video understanding.
\end{IEEEbiography}







\end{document}